\pdfoutput=1

\documentclass[11pt]{article}

\usepackage{EMNLP2022}

\usepackage{times}
\usepackage{latexsym}
\usepackage{enumitem}
\usepackage{comment}

\usepackage[T1]{fontenc}

\usepackage[utf8]{inputenc}

\usepackage{microtype}

\usepackage{inconsolata}
\usepackage{todonotes}
\usepackage{graphicx}
\usepackage{multirow}
\usepackage{booktabs}
\usepackage{arydshln}

\usepackage{amsmath,amsfonts,bm}
\usepackage[nameinlink]{cleveref}

\crefformat{section}{\S#2#1#3} 
\crefname{algorithm}{Alg.}{Algs.}
\crefformat{subsection}{\S#2#1#3}
\Crefname{equation}{Eq.}{Eqs.}
\Crefname{figure}{Fig.}{Figs.}

\makeatletter   
\newcommand{\sveryshortarrow}[1][3pt]{\mathrel{%
    \vcenter{\hbox{\rule[-.5\fontdimen8\scriptfont3]
               {\scriptratio\dimexpr#1\relax}{\fontdimen8\scriptfont3}}}%
   \mkern-4mu\hbox{\let\f@size\sf@size\usefont{U}{lasy}{m}{n}\symbol{41}}}}
\makeatother

\def\eqref#1{equation~\ref{#1}}

\def\1{\bm{1}}

\def\vx{{\bm{x}}}
\def\vy{{\bm{y}}}
\def\vz{{\bm{z}}}

\def\m1{{\bm{1}}}

\DeclareMathAlphabet{\mathsfit}{\encodingdefault}{\sfdefault}{m}{sl}
\SetMathAlphabet{\mathsfit}{bold}{\encodingdefault}{\sfdefault}{bx}{n}

\def\gL{{\mathcal{L}}}
\def\gM{{\mathcal{M}}}

\def\sC{{\mathbb{C}}}

\title{Towards Summary Candidates Fusion}

\author{Mathieu Ravaut$^\clubsuit$$^\diamondsuit$,
Shafiq Joty$^\clubsuit$$^\spadesuit$,
Nancy F. Chen$^\diamondsuit$ \\
$^\clubsuit$ Nanyang Technological University, Singapore\\
$^\diamondsuit$ Institute of Infocomm Research (I$^{2}$R), A$^{*}$STAR, Singapore\\
$^\spadesuit$ Salesforce Research\\
\texttt{\{mathieuj001@e.ntu, srjoty@ntu\}.edu.sg}\\
\texttt{nfychen@i2r.a-star.edu.sg}
}

\begin{document}

\maketitle

\begin{abstract}
Sequence-to-sequence deep neural models fine-tuned for abstractive summarization can achieve great performance on datasets with enough human annotations. Yet, it has been shown that they have not reached their full potential, with a wide gap between the top beam search output and the \emph{oracle} beam. Recently, re-ranking methods have been proposed, to learn to select a better summary candidate. However, such methods are limited by the summary quality  aspects captured by the first-stage candidates. To bypass this limitation, we propose a new paradigm in second-stage abstractive summarization called SummaFusion that fuses several summary candiates to produce a novel abstractive \emph{second-stage} summary. Our method works well on several summarization datasets, improving both the ROUGE scores and qualitative properties of fused summaries. It is especially good when the candidates to fuse are worse, such as in the few-shot setup where we set a new state-of-the-art. We will make our code and checkpoints available at \url{https://github.com/ntunlp/SummaFusion/}.
\end{abstract}

\section{Introduction}
\label{sec:intro}

Leading abstractive summarization methods typically rely on transfer learning and the \emph{pre-train-then-finetune} paradigm. In this approach, a deep sequence-to-sequence neural model is first pre-trained on a very large text corpus, either with a general-purpose text generation objective like \emph{masked span generation} as in T5 \cite{raffel2019exploring}, BART \cite{lewis-etal-2020-bart} and ProphetNet \cite{qi-etal-2021-prophetnet}, or with a pre-training objective specific to summarization as in PEGASUS \cite{zhang2020pegasus} and TED \cite{yang-etal-2020-ted}. Then, the model is fine-tuned on the downstream summarization dataset(s) of interest, which can have a wildly varying amount of human labels, from a few thousands to hundreds of thousands.

\begin{table}[h]
\resizebox{0.98\columnwidth}{!}{
\begin{tabular}{l}
\toprule
\textbf{Source document}:\\ 
\parbox{2.0\columnwidth}{\large this happen yesterday afternoon. i been trying to dual boot mac os and windows on my 
wife's macbook pro. it's a late 2011 model so support from apple is almost nonexistent
which is great when they wanted to charge me chat with them. i convinced them to a 
free chat and learned that apparently my hardware is to out dated to have boot camp 
make a bootable usb. boot camp assistant on this macbook only does cd iso img. \\
...\\ 
3: delete the wrong hard dive part and corrupt the hard drive and have to re format 
the whole computer and lose every file that was saved on her computer since 2011. 
i got yelled at for a good hour. i knew it was my fault but at the same time... 
how in the world have you not backed your things in 4 years!} \\
\midrule
\textbf{Summary candidates (PEGASUS with diverse beam search)}:\\ 
\parbox{2.0\columnwidth}{\large 1: got yelled at for 4 years.\\ 
2: i tried to back up my wife's files on her macbook pro and ended up deleting 4 years worth of data.\\ 
3: tried to back up my wife's files on her macbook pro and ended up deleting 4 years worth of data.\\ 
4: got yelled at by a bunch of people because i backed up my wife's files. \\ 
5 i was trying to back up my wife's files on her macbook pro when i accidentally backed it up.\\ 
6: my wife has been using a late 2011 model for 4 years. \\ 
7: 04/21/2019: fired because i backed up my wife's files on her computer since 2011\\ 
8: decided to make a bootable usb on my windows desktop and accidentally backed up my wife's files for 4 years.\\ 
9 got yelled at for 4 hours by a bunch of people because i backed up my wife's files.\\ 
10: got yelled at for 4 years. went to town to get rid of any doubt.\\ 
11: i backed up my wife's files on her macbook pro and now she's going to lose all her files.\\ 
12: i tried to back up my wife's files but ended up deleting 4 years worth of data.\\ 
13: truck driver: i backed up my wife's data for 4 years.\\ 
14: tried to back up my wife's files and ended up deleting her entire computer.\\ 
15: fired because i backed up my wife's data without realizing it. }\\
\midrule
\textbf{SummaFusion summary}:\\ 
\parbox{2.0\columnwidth}{\large \underline{tried} to \emph{dual boot} my wife's macbook pro \emph{with boot camp assistant} and \underline{ended} up \underline{deleting} 4 years \underline{worth} of \underline{data}.} \\
\midrule
\textbf{Ground truth summary}:\\ 
\parbox{2.0\columnwidth}{\large tried to install windows on macbook and ended up erasing everything without backing up 
and losing 4 years of my wife's work.}\\
\bottomrule
\end{tabular}
}
\caption{\textbf{Qualitative sample from the Reddit TIFU dataset.} Words in the summary from our SummaFusion model which are not in the source document are underlined, and those which are not among any of the first-stage candidates are in italic.}
\label{tab:1}
\end{table}

Sequence-to-sequence models are typically fine-tuned for generation tasks such as summarization with maximum likelihood estimation (MLE): the model is taught to predict only the ground-truth summary given the source, while all {other potential good summary alternatives are not considered.}  This is not ideal since for a subjective task like summarization, there can be several, or even \emph{many} satisfying outputs. 
Besides, at training time, teacher forcing is used \cite{williams1989learning} where the decoder is conditioned on the previous ground truth tokens, while at inference, the model predicts the output sequence auto-regressively by conditioning on its own previous outputs. Once again, this procedure is not ideal as training and inference present a  discrepancy known as \emph{exposure bias} \cite{Bengio-NIPS-15}. Moreover, generation of the most probable sequence becomes intractable due to a large vocabulary size, and typically a decoding method is used to approximate the best summary. Beam search \cite{reddy1977speech} has been a common choice for decoding, but other methods such as nucleus sampling \cite{holtzman2019curious} are gaining traction as potential alternatives, usually with a focus on diversity in the generation. 

When decoding the summary, the decoding method keeps track of several hypotheses, before outputting a single one and discarding the others. An \emph{oracle} is defined as the summary candidate from the pool of hypotheses which maximizes the metric of choice (e.g. {\sc{Rouge}} \cite{lin-2004-rouge}) when evaluated against the target. As observed by several recent studies \cite{liu-liu-2021-simcls,ravaut-etal-2022-summareranker}, the discrepancies between training and inference together with the approximate decoding lead to models not being utilized to their full capacity. As a consequence, there is a wide gap between the top candidate and the oracle performance. For instance, \cite{ravaut-etal-2022-summareranker} report a 10.07 {\sc{Rouge-1}} points difference between the top beam and the diverse beam search oracle on CNN/DM. This motivates second-stage methods to learn to select a better candidate with having access to a more global context, free from the auto-regressive constraint which restricts access to only previous context. Summarizing in multiple stages is arguably also closer to how humans compose a summary \cite{cao-etal-2018-retrieve}.

Existing second-stage summarization methods design a training objective to improve candidate selection among first-stage candidates, either through a new model \cite{ravaut-etal-2022-summareranker}, or re-using the first-stage model \cite{liu-etal-2022-brio}. However, sticking to first-stage candidates may not be ideal as they are bounded by the quality of the first-stage model. Despite the oracle average results being high, its variance is very high too (see \Cref{sec:appendix_a} \Cref{tab:a_1_2}), and in some cases all candidates are sub-optimal. Alternative decoding methods to beam search, while generating more diverse summaries, do not solve the issue as high diversity among candidates usually means loss in performance in the output candidate \cite{narayan-etal-2022-well}. 

To bypass these limitations, in this work we propose \emph{SummaFusion}, an \emph{abstractive} second-stage summarization model. Re-using both the source and first-stage summary candidates, SummaFusion generates a new summary from scratch, in an abstractive manner. It produces summary candidates which are closer to the ground-truth ones, resulting in relative {\sc{Rouge}} improvements of up to 17.98\% across three very abstractive benchmarks. The model is flexible and can adjust to varying number of summary candidates. Besides, fused summaries present several interesting properties such as being abstractive with regards to both the source and the set of first-stage candidates, more fluent and more factually consistent. It performs well especially on lower-quality pools of summary candidates, where one needs a second-stage summarizer the most. For instance, this happens often in few-shot scenarios, where the first-stage model generally lacks enough supervision in order to learn to produce good summaries.  To get a glimpse of our model behavior, we refer the reader to the example in \Cref{tab:1}, where the fused summary is much better than all candidates. 


Our contributions in this paper are the following:
\begin{itemize}[leftmargin=*,topsep=2pt,itemsep=2pt,parsep=0pt]
    \item We introduce summary candidates fusion, a novel approach to second-stage summarization. 
    \item We demonstrate the fixing behavior of our SummaFusion: it is designed for very abstractive summarization tasks, and works better on difficult data points for the base model, as well as in few-shot setups. In these cases, it dramatically drives up the {\sc{Rouge}} of the base model. 
    \item We conduct a thorough qualitative analysis, and assess that SummaFusion indeed generates summaries deemed better according to humans. 
\end{itemize}

\section{Related Work}
\label{sec:related}

Second-stage methods have recently enabled strong progress in the state-of-the-art of abstractive summarization research. GSum \cite{dou-etal-2021-gsum} uses additional discrete guidance signal such as salient sentences predicted by an extractive model to better guide the abstractive system. While abstractive summarization models are trained to maximize MLE at the token-level, second-stage methods usually work at the sequence-level. ConSum \cite{sun2021alleviating} and SeqCo \cite{xu2021sequence} fine-tune the model with a different, contrastive loss to assign more confidence to higher-quality summary candidates. RefSum \cite{liu-etal-2021-refsum} uses meta-learning to re-rank summary candidates produced by several base systems. SummaReranker \cite{ravaut-etal-2022-summareranker} and SimCLS \cite{liu-liu-2021-simcls} train a RoBERTa to re-rank candidates, the former with multi-label binary cross-entopy, the latter with contrastive learning and a ranking loss. BRIO \cite{liu-etal-2022-brio} re-uses the base model for a second-round of fine-tuning with both the cross-entropy loss and a candidate-level ranking loss.

Existing fusion work in summarization focuses on sentence fusion. Fusing several sentences for the purpose of summarization was first proposed by \citet{barzilay-mckeown-2005-sentence}, paving the way for more abstractive summaries. \citet{weiss2021extending} later proposed a much larger dataset for sentence fusion in multi-document abstractive summarization, driving up model performance. Through a thorough human evaluation, \citet{lebanoff-etal-2019-analyzing} ask annotators to label which type of fusion of sentences is taking place while also rating the sentence properties for sentences generated by several abstractive systems. In a follow-up work \cite{lebanoff-etal-2020-learning}, the authors build a Transformer model \cite{vaswani2017attention} enriched with sentence structure information for the explicit goal of fusing sentences, and evaluate the model on a dataset dedicated to sentence fusion. The same authors also introduce a cascade model \cite{lebanoff-etal-2020-cascade} for abstractive summarization which contains a fusion mechanism based on combining highlighted phrases of the source text.

 To the best of our knowledge, there is no method yet to fuse or combine in an abstractive manner \emph{entire} summary candidates (not just sentences).

\section{Model}
\label{sec:model}

\begin{figure}[t!]
    \centering 
    \includegraphics[width=\columnwidth]{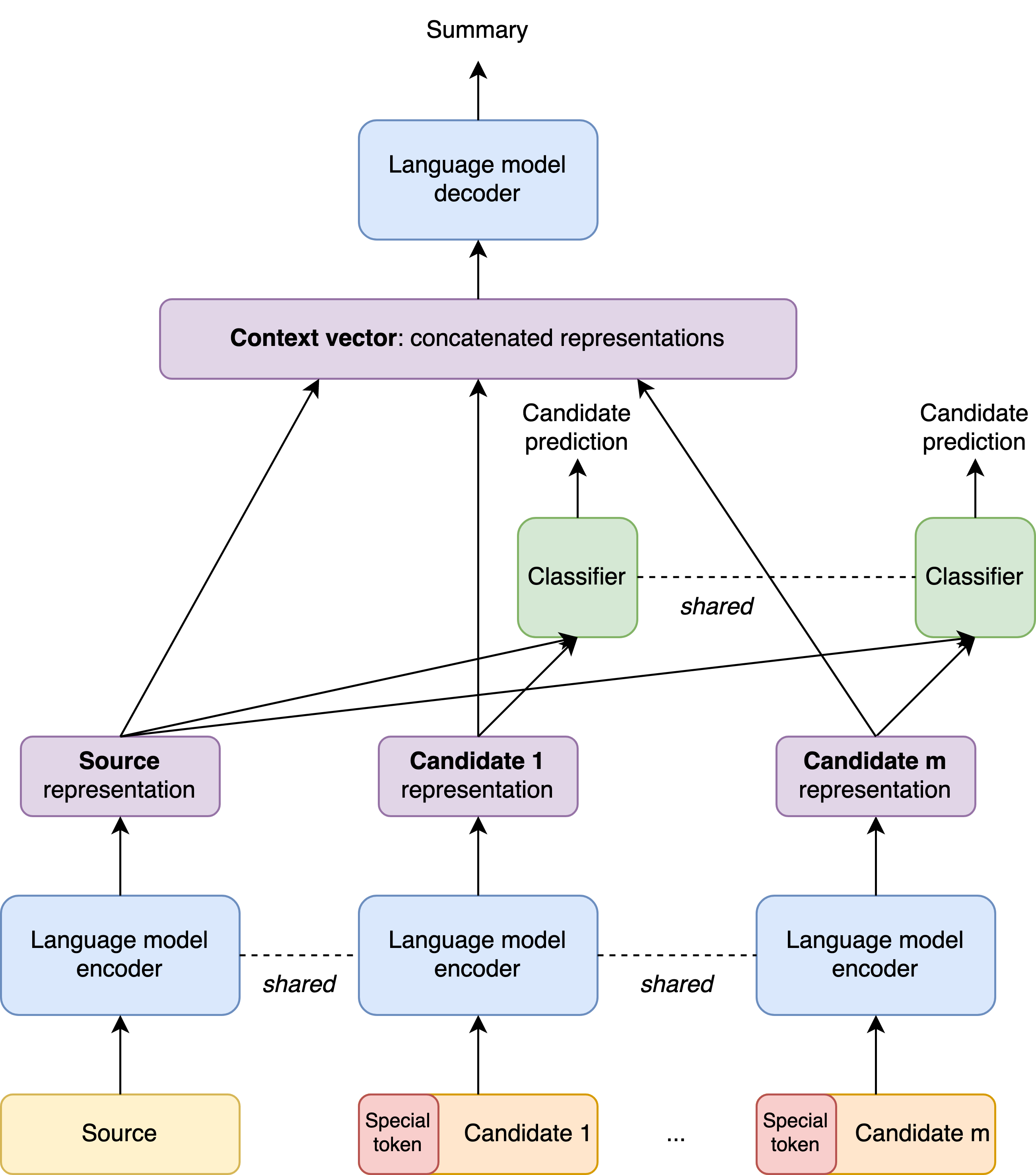}
    \caption{\textbf{SummaFusion model architecture.} SummaFusion encodes the source and each of the $m$ summary candidates separately, then concatenate their representations on the sequence dimension before decoding.}
    \label{fig:1}
\end{figure}

If $\vx_i$ is a source document and $\vy_i$ its associated target summary, $B$ is the first stage model generating $m$ candidates $\sC_i = \{C_1, \ldots, C_m\}$, the second-stage fusion model $\theta$ is trained to maximize the likelihood of the target given $\vx_i$ \emph{and} $\sC_i$: 

\begin{equation}
    \small 
    \hat{\theta} = \arg \max_{\theta} \log  p_{\theta}(\vy_i | \vx_i, \sC_i) 
\end{equation}
 
where the joint distribution over the target tokens $p(\vy_i | \vx_i, \sC_i; \theta)$ is modeled as auto-regressive generation with the following cross-entropy loss for a target summary $\vy_{i} = (y_{1}, \ldots, y_{l})$ of length $l$:

\begin{equation}
 \small 
     \gL_{\text{gen}} = - \sum_{j=1}^{l} \log  p_{\theta} (y_{j} | y_{1}, \ldots, y_{j-1}, \vx_i, \sC_i)
\end{equation}
This formulation is essentially the same as the one typically used to train the base model $B$: auto-regressive cross-entropy loss with teacher forcing. The only difference is that in the second stage, the model can also condition on the first stage candidates $\sC_i$ on top of the source $\vx_i$. \Cref{fig:1} shows the overall architecture of our fusion model.


\subsection{Fusing Source and Summary Candidates}

Knowing that among the pool of first-stage summary candidates, some are of high quality, we give the fusion model access to the entire set $\sC_i$. At the same time, to enable the model to deviate from the candidates if needed, we also condition on the source $\vx_i$. We first encode the source $\vx_i$ and each candidate $C_{k} \in \sC_i$ separately with the encoder:

\begin{equation}
    \small
    \vz_{\vx_{i}} = \theta_{\text{enc}}(\vx_i); \hspace{2em} \vz_{C_{k}} = \theta_{\text{enc}}(C_{k})
    \label{eq:1}
\end{equation}

We then concatenate all these token-level representations from the encoder on the \emph{sequence} dimension, resulting in a unique, long context vector:

\begin{equation}
    \small
    \vz_{i} = \text{concat}(\vz_{\vx_{i}}, \vz_{C_{1}}, \ldots, \vz_{C_{m}})
\end{equation}

Finally, the decoder performs cross-attention on $\vz_{i}$ to generate the output token probabilities:

\begin{equation}
    \small
    p_{\theta}(\vy_i | \vx_i, \sC_i) = \theta_{\text{dec}}(\vz_{i})
\end{equation}

\begin{table*}
\resizebox{\textwidth}{!}{
\begin{tabular}{llccccccccccc}
\toprule
\multirow{2}{*}{\textbf{Dataset}} 
& \multirow{2}{*}{\textbf{Domain}} 
& \multicolumn{3}{c}{\textbf{\# Data points}} 
& \multicolumn{2}{c}{\textbf{\# Words}} 
& \multicolumn{2}{c}{\textbf{\# Tokens}} 
& \multirow{2}{*}{\textbf{\begin{tabular}[c]{@{}c@{}}\textbf{Compression} \\ \textbf{ratio (\%)}\end{tabular}}} 
& \multicolumn{3}{c}{\textbf{Abstractiveness (\%)}}  \\
& & Train & Val & Test 
& Doc. & Summ. 
& Doc. & Summ. 
& & 1-grams & 2-grams & 3-grams \\
\midrule
XSum & News & 204045 & 11332 & 11334 & 414.51 & 22.96 & 456.96 & 26.01 & 5.27 & 33.98 & 83.33 & 95.52 \\
Reddit TIFU & Social media & 33704 & 4213 & 4222 & 385.59 & 20.59 & 466.44 & 25.99 & 6.55 & 28.78 & 77.63 & 92.73 \\
SAMSum & Dialogue & 14732 & 818 & 819 & 123.72 & 23.39 & 133.07 & 25.66 & 23.77 & 33.88 & 79.02 & 90.10 \\
\bottomrule
\end{tabular}
}
\caption{\textbf{Statistics} on the datasets that we used for experiments. Tokens counts are calculated based on PEGASUS tokenization. Compression ratio is defined as the ratio between the number of sentences in the summary and the number of sentences in the source.}
\label{tab:2}
\vspace{-1.0em}
\end{table*}

Our approach of concatenating after encoding is inspired by the Fusion-in-Decoder method \cite{izacard-grave-2021-leveraging}, which is more suited to our problem than concatenating before encoding. Indeed, if we note $n$ the source length and $l$ the summary length, the complexity of the self-attention when concatenating before would be: $\mathcal{O}((n + m.l)^{2} + n.l + l^{2})$, while with our approach, it becomes:
$\mathcal{O}(n^{2 }+ m.l^{2} + (n + m.l).l + l^{2})$. Knowing that in summarization, we have $l << n$, concatenating after is less computationally expensive. Besides, self-attention between summary candidates does not yield any additional value in our problem while being computationally expensive.

\subsection{Candidate-level Information}

Summary candidates $C_1, \ldots, C_m$ are initially ordered following their diverse beam search order (by group, and then log-probability within each group). To enrich the model with this ranking information, we append a special token $[C_{k}]$ in front of the k-th candidate $C_k$. When concatenating the representations, this also gives the model information on where each summary candidate representation starts. 

To further make the model aware of the quality of each summary candidate, we also add a classification component. Given a candidate $C_k$ and a summary evaluation metric $\mu$ (e.g. {\sc{Rouge}} \cite{lin-2004-rouge}), the model has to predict whether $C_k$ is maximizing $\mu$ among the summary candidates in $\sC_i$. We frame this as a binary classification problem and the associated binary cross-entropy loss is:

\begin{equation}
    \small 
    \begin{split}
    \gL_{\text{cls}}^{\mu} = \sum_{k=1}^{m} &- z_k^{\mu} \log p_{\theta}(C_k) \\ &- (1-z_k^{\mu}) \log (1 - p_{\theta}(C_{k}))
\end{split}
\end{equation}
 
where $z_k^{\mu}$ is $1$ if the candidate $C_k$ maximizes the metric $\mu$, $0$ otherwise, and $p_{\theta}(C_k)$ is the probability predicted by the model that the candidate is positive. Following SummaReranker \cite{ravaut-etal-2022-summareranker}, we use a multi-label approach and train the classification model jointly for $\gM = \{\mu_1, \ldots, \mu_M\}$ several metrics, and we also condition on the source representation as input to the classifier (concatenated with the candidate representation). The final classification loss is:

\begin{equation}
    \small 
    \gL_{\text{cls}} = \frac{1}{M}\sum_{\mu_m \in \gM} \gL_{\text{cls}}^{\mu_m}
\end{equation}

In practice, we use metrics $\gM = \{${\sc{Rouge-1}}, {\sc{Rouge-2}}, {\sc{Rouge-L}}$\}$.

Our final model loss is a combination of both the generation and classification losses:

\begin{equation}
    \small 
    \gL = \gL_{\text{gen}} + \lambda. \gL_{\text{cls}}
\end{equation}

where $\lambda$ is hyper-parameter to be tuned on the validation set.

\subsection{Input Dropout}
\label{sec:3_3}

Our model has access to several information streams at once. To make it robust and not only learn to rely on a single channel (for instance, only the source document), we use input dropout \cite{provilkov-etal-2020-bpe}. We use two variants of input dropout during training as follows:

\begin{itemize}[leftmargin=*]
    \item \textbf{Source dropout}: To prevent the model from solely relying on source, we replace the source $\vx_i$ with a placeholder token with probability $p_{\text{src}}$.  

    \item \textbf{Candidates dropout}: We subsample with uniform probability $k \in \{2, \ldots, m\}$ summary candidates to keep, replacing the other $m-k$ by placeholder tokens. This stops the model from only copying the summary candidate from a fixed position (for instance, the top beam). 

\end{itemize}

\section{Experimental Settings}
\label{sec:experiments}

\begin{table*}[t!]
\resizebox{\textwidth}{!}{
\begin{tabular}{lcc|cccc|cccc|cccc}
\toprule
& & 
& \multicolumn{4}{c}{\textbf{XSum}} 
& \multicolumn{4}{c}{\textbf{Reddit TIFU}}                                       
& \multicolumn{4}{c}{\textbf{SAMSum}} \\
\textbf{Model}                   
& \textbf{Stage} 
& \textbf{Candidates} 
& \textbf{R-1} & \textbf{R-2} & \textbf{R-L} & \begin{tabular}[c]{@{}c@{}}\textbf{Gain} \\ \textbf{(\%)}\end{tabular} 
& \textbf{R-1} & \textbf{R-2} & \textbf{R-L} & \begin{tabular}[c]{@{}c@{}}\textbf{Gain} \\ \textbf{(\%)}\end{tabular} 
& \textbf{R-1} & \textbf{R-2} & \textbf{R-L} & \begin{tabular}[c]{@{}c@{}}\textbf{Gain} \\ \textbf{(\%)}\end{tabular}  \\
\midrule
PEGASUS (\textbf{BS}) \cite{zhang2020pegasus} & 1 & 8 & 47.21 & 24.56 & 39.25 & \_ & \emph{26.63} & \emph{9.01} & \emph{21.60} & \_ & \_ & \_ & \_ & \_ \\
PEGASUS (ours, \textbf{BS}) & 1 & 15 & 47.33 & 24.75 & 39.43 & \_ & 26.28 & 9.01 & 21.52 & \_ & 52.04 & 27.53 & 43.54 & \_ \\
PEGASUS (ours, \textbf{DBS}) & 1 & 15 & 46.78 & 23.77 & 38.70 & \_ & 25.67 & 8.07 & 20.97 & \_ & 51.35 & 26.89 & 42.65 & \_ \\
PEGASUS (ours, \textbf{DBS}) - random & 1 & 15 & 42.95 & 19.64 & 34.14 & -11.47 & 23.63 & 6.58 & 19.07 & -9.94 & 46.49 & 19.60 & 41.44 & -11.06 \\
PEGASUS (ours, \textbf{DBS}) - \emph{oracle} & 1 & 15 & \emph{56.76} & \emph{34.46} & \emph{50.18} & \emph{29.42} & \emph{35.94} & \emph{14.42} & \emph{30.24} & \emph{47.30} & \emph{62.74} & \emph{39.07} & \emph{58.54} & \emph{32.63} \\
\midrule
GSum (BART + MatchSum) \cite{dou-etal-2021-gsum} & 2 & \_ & 45.40 & 21.89 & 36.67 & \_ & \_ & \_ & \_ & \_ & \_ & \_ & \_ & \_ \\
PEGASUS + ConSum \cite{sun2021alleviating} & 2 & \_ & 47.34 & 24.67 & 39.40 & \_ & \_ & \_ & \_ & \_ & \_ & \_ & \_ & \_ \\
BART + SeqCo \cite{xu2021sequence} & 2 & \_ & 45.65 & 22.41 & 37.04 & \_ & \_ & \_ & \_ & \_ & \_ & \_ & \_ & \_ \\
BART (\textbf{DBS}) + SimCLS \cite{liu-liu-2021-simcls} & 2 & 16 & 47.61 & 24.57 & 39.44 & \_ & \_ & \_ & \_ & \_ & \_ & \_ & \_ & \_ \\
PEGASUS (\textbf{BS}) + SummaReranker* \cite{ravaut-etal-2022-summareranker} & 2 & 15 & 48.12 & 24.95 & 40.00 & \_ & 29.57 & 9.70 & 23.29 & \_ & \textbf{52.97} & 27.18 & 43.82 & \_ \\
PEGASUS (\textbf{DBS}) + SummaReranker* \cite{ravaut-etal-2022-summareranker} & 2 & 15 & 47.04 & 23.27 & 38.55  & -0.37 & 28.71 & 8.73 & 22.79 & 10.07 & 52.05 & 26.17 & 42.57 & -0.07 \\
BART (\textbf{DBS}) + BRIO \cite{liu-etal-2022-brio} & 2 & 16 & \textbf{49.07} & \textbf{25.59} & \textbf{40.40} & \_ & \_ & \_ & \_ & \_ & \_ & \_ & \_ & \_ \\
\hdashline
PEGASUS (ours, \textbf{DBS}) + \textbf{SummaFusion-base} & 2 & 15 & 46.16 & 23.55 & 38.53 & -0.93 & 27.52 & 9.01 & 22.23 & 7.40 & 51.61 & 26.53 & 43.09 & 0.27 \\
PEGASUS (ours, \textbf{DBS}) + \textbf{SummaFusion-large} & 2 & 15 & 47.08 & 24.05 & 38.82 & 0.63 & \textbf{30.08} & \textbf{10.48} & \textbf{23.99} & 17.98 & 52.76 & \textbf{28.24} & \textbf{43.98} & 3.37 \\
\bottomrule
\end{tabular}
}
\caption{\textbf{{\sc{Rouge}} results} on the three datasets with PEGASUS base model. The first block shows performance from generated summaries after the first stage, while the second block corresponds to second-stage summarization models. \textbf{BS} denotes beam search, \textbf{DBS} is diverse beam search, and \textbf{R-1/2/L} means {\sc{Rouge-1/2/L}}. \textbf{Gain} is the relative gain over the mean of {\sc{Rouge-1}}, {\sc{Rouge-2}} and {\sc{Rouge-L}} \emph{from our own PEGASUS DBS baseline}. * SummaReranker is trained on its recommended setup of a mix of beam search and diverse beam search summary candidates. Results in italic are not directly comparable, as they either involve accessing the target (oracle), or are obtained on a different split (Reddit).} 
\label{tab:3}
\vspace{-1.0em}
\end{table*}

\subsection{Abstractive Summarization Tasks}

We apply our SummaFusion to three popular abstractive summarization datasets, each covering a different domain. Datasets were chosen for their high level of abstraction, and are as follows:

\begin{itemize}[leftmargin=*]
    \item \textbf{XSum} \cite{narayan-etal-2018-dont} is the task of extreme summarization. It consists in news articles being compressed into highly-abstractive, single-sentence summaries. The dataset spans 227k articles from the BBC from 2010 to 2017.
    
    \item \textbf{Reddit TIFU} \cite{kim-etal-2019-abstractive} corresponds to real-life stories written in the form of long blogs on the popular Reddit social media. It is made of 120k posts. As in other summarization papers \cite{zhang2020pegasus, ravaut-etal-2022-summareranker}, we use the  TIFU-long subset, containing 37k posts. 
    
    \item \textbf{SAMSum} \cite{gliwa-etal-2019-samsum} is a dialogue summarization dataset containing 17k conversations. Compression ratio is significantly higher on this dataset, as the source conversations are short.
\end{itemize}

We excluded the popular CNN/DM dataset since it is highly extractive \cite{hermann2015teaching, see-etal-2017-get}. Detailed statistics on all our datasets can be found in \Cref{tab:2}. As seen, on each dataset, there is a high proportion of n-grams in summaries which are not found in the source, highlighting the very abstractive nature of these summarization tasks. To download datasets, we use HuggingFace \emph{datasets} library \cite{lhoest-etal-2021-datasets}.

\subsection{Model Details}

As base model $B$, we use PEGASUS \cite{zhang2020pegasus}, a strong baseline on our selected datasets. To generate candidates, we use diverse beam search \cite{vijayakumar2016diverse} with 15 beams, following \cite{ravaut-etal-2022-summareranker}. Diverse beam search is much more suited than beam search for our setup, due to the greater variety among the whole pool of candidates and consequently higher oracle performance (see \Cref{sec:appendix_a} for oracle results). 

For SummaFusion encoder and decoder, we use BART \cite{lewis-etal-2020-bart} and experiment with both the \emph{base} and the \emph{large} versions, referred to as \emph{SummaFusion-base} and \emph{SummaFusion-large}. Although re-using PEGASUS is technically feasible, we found that SummaFusion benefits from diverse models. We train SummaFusion in both full-shot and three few-shot setups (10-shot, 100-shot, and 1000-shot).


\begin{figure*}[t]
    \centering 
    
    \minipage{0.33\textwidth}
      \includegraphics[width=\linewidth]{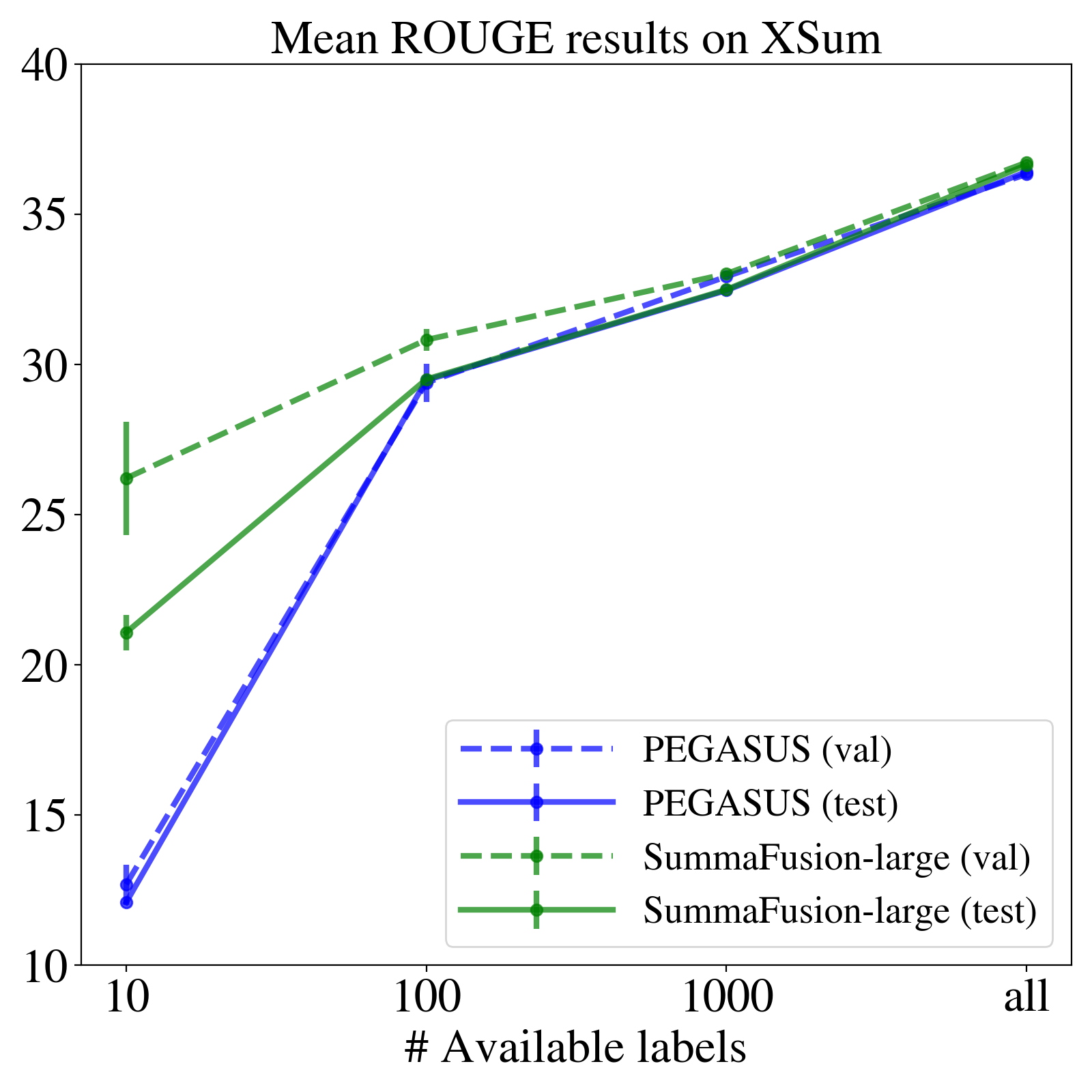}
    \endminipage\hfill
    \minipage{0.33\textwidth}
      \includegraphics[width=\linewidth]{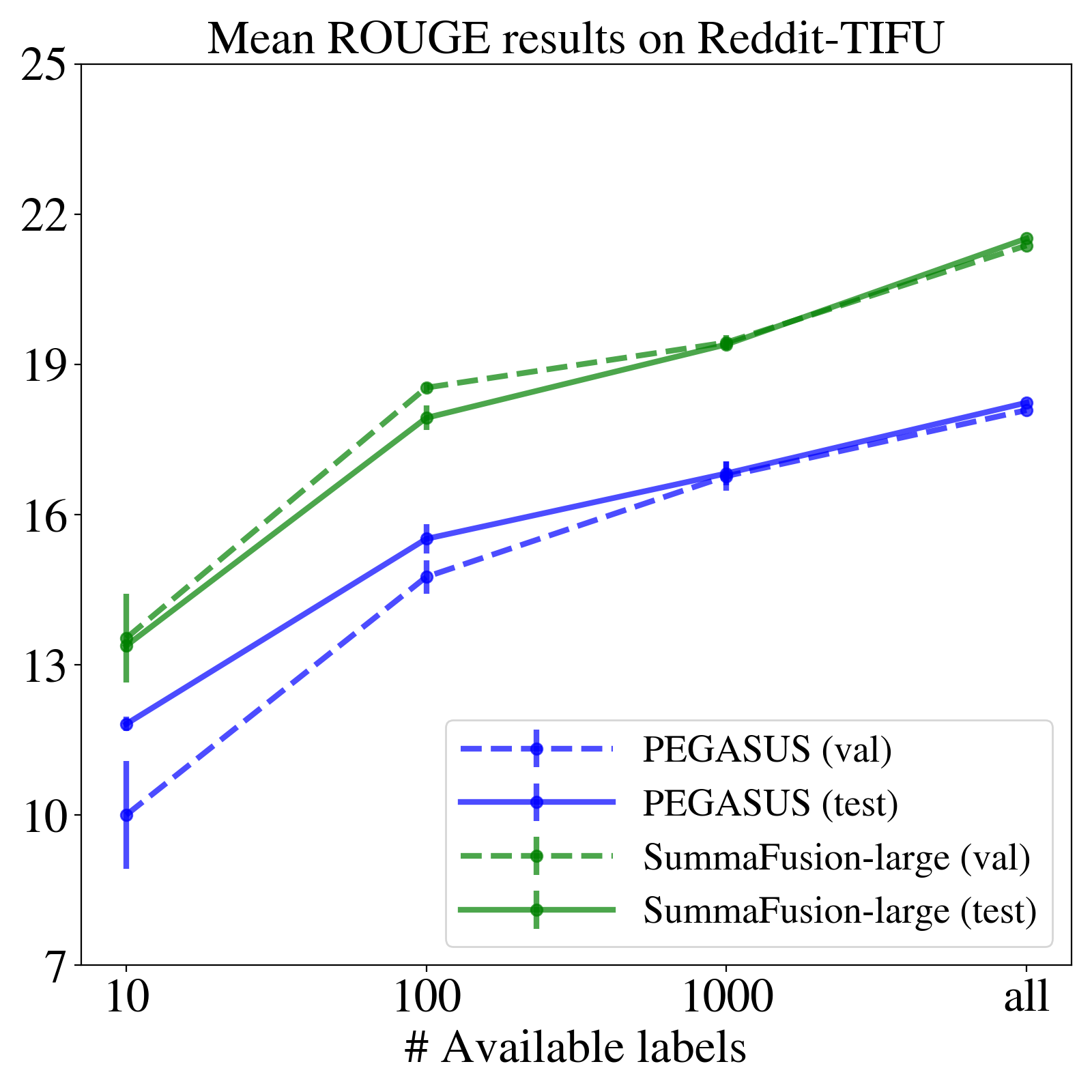}
    \endminipage\hfill
    \minipage{0.33\textwidth}%
      \includegraphics[width=\linewidth]{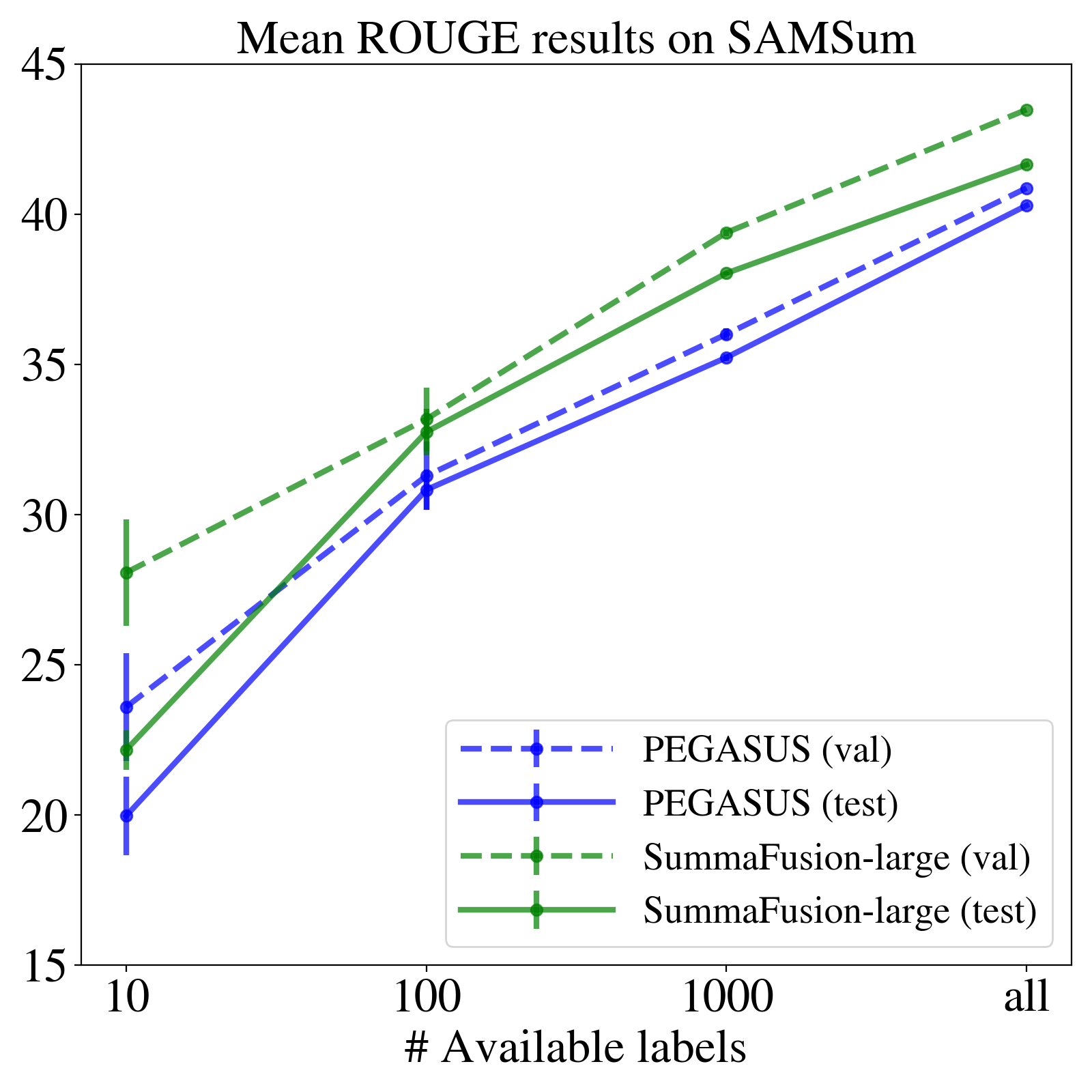}
    \endminipage
    
    \caption{\textbf{Few-shot {\sc{Rouge}} results} on the three datasets. Vertical bars on the dots represent standard deviation over the random seeds. Results on "all" available correspond to full-dataset fine-tuning (see \Cref{tab:3}). Dashed lines correspond to validation results, and full lines to test set results.}
    \label{fig:2}
\end{figure*}

\subsection{Training and Optimization}

As a second-stage supervised method, SummaFusion suffers from the inherent train-test distribution mismatch. This means that one cannot train SummaFusion on outputs of the base model $B$ on the training set, as the generated summaries would be of a different distribution than the generated summaries on the validation and test sets.\footnote{Since $B$ is trained on the same training set, the generated summaries on this set are expected to be of higher quality than the ones on the unseen validation/test sets.} To alleviate this issue, we follow the 50-50 split approach used in SummaReranker \cite{ravaut-etal-2022-summareranker}. We split each training set in two equal halves, fine-tune $B$ on each half and infer it on the other half, then train the fusion model on the concatenation of both inferred halves. At inference, we use the \emph{transfer} approach and apply SummaFusion on candidates generated by another base model fine-tuned on the \emph{entire} training set.

We follow XSum and SAMSum provided train:val:test splits, and use the same 80:10:10 split as \cite{ravaut-etal-2022-summareranker} on Reddit TIFU. On XSum, we use fine-tuned PEGASUS checkpoints hosted by HuggingFace \emph{transformers} library \cite{wolf-etal-2020-transformers}. On the other two datasets, we fine-tune our own PEGASUS, starting with the pre-trained checkpoint shared on \emph{transformers}. Hyper-parameters used for fine-tuning the base PEGASUS can be found in the \Cref{sec:appendix_b} \Cref{tab:a_2_1}, and \Cref{tab:a_2_2} for summary generation hyper-parameters. 

We initialize SummaFusion backbone BART with the pre-trained checkpoint from \cite{wolf-etal-2020-transformers}. To optimize the model, we train for 5 epochs with Adam optimizer \cite{kingma2014adam} and a constant learning rate of 2e-5. We found $\lambda=1.0$ to work well and used it in all the results we report. We generate SummaFusion summaries with beam search using beam width 10. Detailed SummaFusion fine-tuning hyper-parameters are in \Cref{sec:appendix_c}.

In the few-shot setups, we use three random seeds, and for each seed sample randomly a training set and a validation set each of the corresponding few-shot size. We show validation and test results averaged over the three few-shot models, alongside corresponding standard deviations.

\section{Evaluation}
\label{sec:evaluation}


We compare SummaFusion outputs with our own PEGASUS with 15 diverse beams baseline (\emph{PEGASUS (ours)}), as well as \emph{PEGASUS-random}, a baseline consisting in randomly selecting a summary candidate. We also include the oracle for reference (\emph{PEGASUS-oracle}) and compare with PEGASUS reported results \cite{zhang2020pegasus}. We use {\sc{Rouge-1}}, {\sc{Rouge-2}}, and {\sc{Rouge-L}} \cite{lin-2004-rouge} and their mean as quantitative metrics to assess summary closeness to the target.

\begin{table}[t!]
\setlength{\tabcolsep}{0.5em}
\resizebox{\columnwidth}{!}{
\begin{tabular}{lcccccc}
\toprule
\multirow{2}{*}{\textbf{Model}} & \multicolumn{3}{c}{\textbf{10-shot}} & \multicolumn{3}{c}{\textbf{100-shot}} \\
 & \textbf{R-1} & \textbf{R-2} & \textbf{R-L} & \textbf{R-1} & \textbf{R-2} & \textbf{R-L} \\
\midrule
\multicolumn{7}{c}{\textit{\textbf{XSum}}} \\
\midrule
PEGASUS \cite{zhang2020pegasus}                     & 19.39 & 3.45 & 14.02 & 39.07 & 16.44 & 31.27 \\
PEGASUS (ours)                                      & 19.38 & 3.31 & 13.60 & \textbf{39.90} & 16.86 & 31.67 \\
PSP \cite{liu2022psp}                               & \_ & \_ & \_ & 32.50 & 10.83 & 25.03 \\
\hdashline
SummaReranker \cite{ravaut-etal-2022-summareranker} & 23.79 & 6.24 & 17.41 & 38.43 & 15.44 & 29.99 \\
SummaFusion-large                                   & \textbf{30.41} & \textbf{9.92} & \textbf{22.93} & 39.86 & \textbf{17.01} & \textbf{31.68} \\
\midrule
\multicolumn{7}{c}{\textit{\textbf{Reddit TIFU}}} \\
\midrule
\emph{PEGASUS} \cite{zhang2020pegasus}              & \emph{15.3}6 & \emph{2.91} & \emph{10.76} & \emph{16.64} & \emph{4.09} & \emph{12.92} \\
PEGASUS (ours)                                      & 18.39 & 3.34 & 13.23 & 22.82 & 5.85 & 17.88 \\
\hdashline
SummaReranker \cite{ravaut-etal-2022-summareranker} & 17.49 & 3.28 & 12.88 & 23.38 & 5.57 & 18.04 \\
SummaFusion-large                                   & \textbf{20.79} & \textbf{4.77} & \textbf{14.58} & \textbf{26.09} & \textbf{7.51} & \textbf{20.22} \\    
\midrule
\multicolumn{7}{c}{\textit{\textbf{SAMSum}}} \\
\midrule
PEGASUS (ours)                                      & 28.47 & 8.59 & 22.87 & 42.09 & 16.85 & 33.54 \\
\hdashline 
SummaReranker \cite{ravaut-etal-2022-summareranker} & 29.72 & 8.83 & 23.61 & 41.27 & 15.18 & 31.94 \\
SummaFusion-large                                   & \textbf{32.00} & \textbf{9.93} & \textbf{24.59} & \textbf{44.41} & \textbf{18.84} & \textbf{35.04} \\    
\bottomrule
\end{tabular}
}
\caption{\textbf{Detailed few-shot {\sc{Rouge}} results}. On Reddit TIFU, PEGASUS results are in italic as they are not directly comparable (different train:val:test split). Both second-stage methods SummaReranker and SummaFusion (bottom blocks) are trained and inferred on the same DBS candidates from our PEGASUS baseline.}
\label{tab:4}
\end{table}

\subsection{Full-shot Results}

Full-shot results with SummaFusion-base and SummaFusion-large for all datasets are displayed in \Cref{tab:3}. SummaFusion-large improves the {\sc{Rouge-1}} compared to the PEGASUS with diverse beam search baseline by 0.30 {\sc{Rouge-1}} points on XSum, 4.41 points on Reddit TIFU and 1.41 points on SAMSum. We notice that SummaFusion helps the most relatively on Reddit TIFU, on the which the baseline performance is sensibly lower. Although not from the same type of technique, SummaFusion is comparable to other recent second-stage methods on XSum \cite{sun2021alleviating}.


\begin{figure*}[t]
    \centering 
    
    \minipage{0.33\textwidth}
      \includegraphics[width=\linewidth]{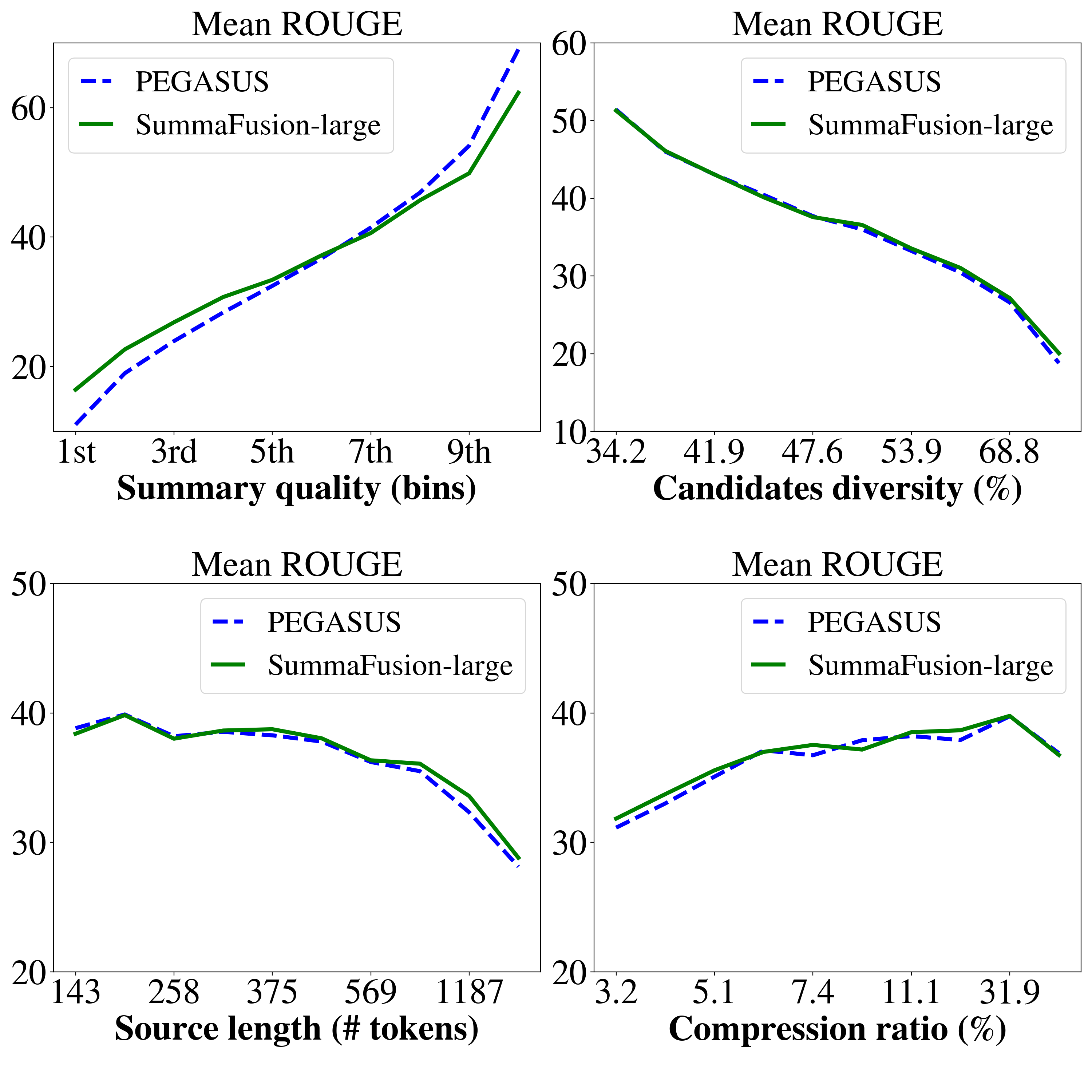}
      \caption*{XSum}
      \label{fig:3_1}
    \endminipage\hfill
    \minipage{0.33\textwidth}
      \includegraphics[width=\linewidth]{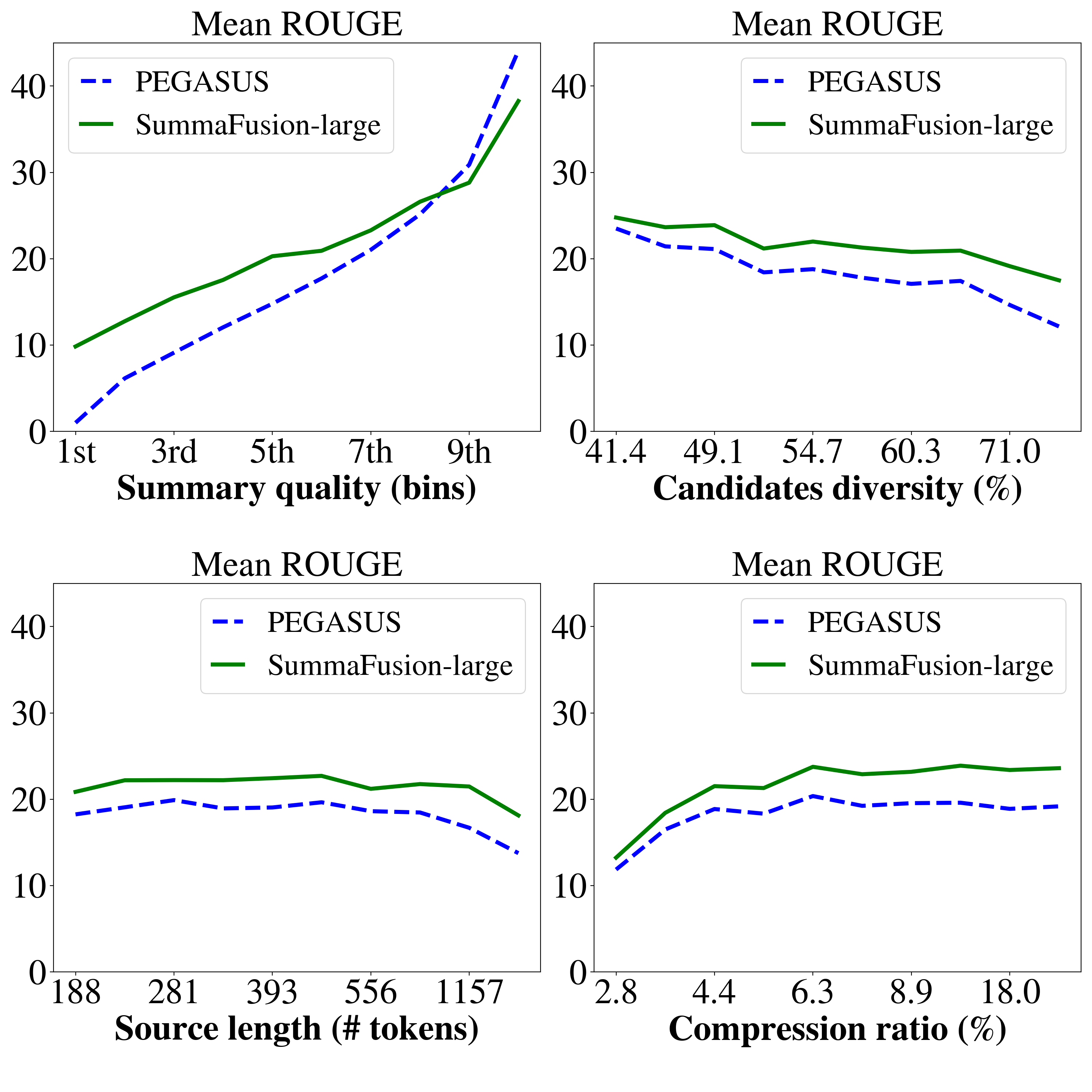}
      \caption*{Reddit-TIFU}
      \label{fig:3_2}
    \endminipage\hfill
    \minipage{0.33\textwidth}%
      \includegraphics[width=\linewidth]{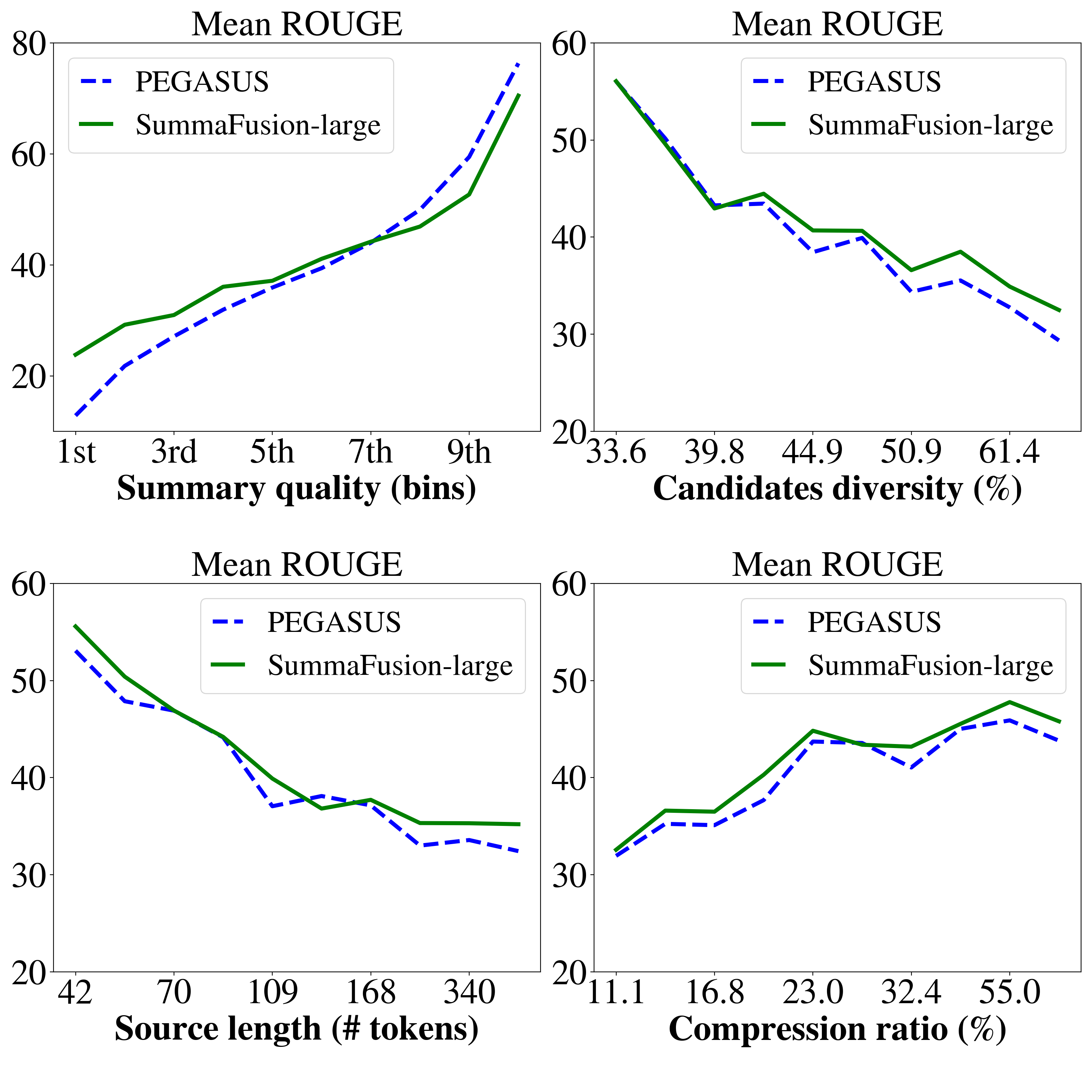}
      \caption*{SAMSum}
      \label{fig:3_3}
    \endminipage
    \caption{\textbf{Fine-grained analysis} on the three datasets. We split data points over four features: average quality of summary candidates pools (top left plots), average diversity of summary candidate pools (top right plots), source length (bottom left plots), and compression ratio (bottom right plots). Across each feature, we split the test set into 10 bins of equal size, and within each bin compute the mean {\sc{Rouge}} of the baseline PEGASUS top beam as well as the mean {\sc{Rouge}} of SummaFusion-large.}
    \label{fig:3}
    \vspace{-1.0em}
\end{figure*}

\subsection{Few-shot Results} 
\label{sec:few-shot}

Next, we apply SummaFusion to three few-shot scenarios: 10, 100 and 1000 labelled data points, respectively. Results are shown in \Cref{fig:2} and \Cref{tab:4}. We notice that with a lower quantity of annotated data, SummaFusion relative gain is higher. Notably, SummaFusion dramatically improves {\sc{Rouge}} on 10-shot summarization. As seen in \Cref{tab:4}, SummaFusion is on par with or better than state-of-the-art few-shot (10-shot and 100-shot) abstractive summarization models, including the comparable second-stage method SummaReranker. PEGASUS remains a very strong 100-shot baseline. We exclude WikiTransfer \cite{fabbri-etal-2021-improving} from the comparison as it was specially designed for few-shot \emph{transfer} and it leverages additional data (Wikipedia) before few-shot fine-tuning.

These results point to a common direction: SummaFusion works better on \emph{lower quality candidates}, "fixing" them into a much better summary. In the following section, we analyze this hypothesis.

\section{Analysis}
\label{sec:analysis}

\subsection{When Is Summary Fusion Helpful?}

The previous section suggested that SummaFusion is better on lower-quality base candidates, such as in Reddit-TIFU or few-shot setups. To verify this hypothesis and better characterize the fixing behavior of SummaFusion, we split the test set across four different features:
\begin{itemize}[leftmargin=*,topsep=2pt,itemsep=2pt,parsep=0pt]
    \item \textbf{Summary quality}: this is the mean {\sc{Rouge}} with the target averaged over all diverse beam search candidates produced by the base model. This feature assesses the overall quality of the initial set of summary candidates.
    
    \item \textbf{Candidates diversity}: we compute $1-\textsc{Rouge-1}$ for all pairs of summary candidates and average the results. Since a high {\sc{Rouge-1}} between candidates indicates that they overlap, the average $1-\textsc{Rouge-1}$ feature measures how \emph{diverse} is the pool of summary candidates.
    
    \item \textbf{Source length}: this corresponds to the number of words in the source document. Modeling long inputs is challenging so we expect summarization models to work less well on longer documents.
    
    \item \textbf{Compression ratio}: this is the ratio between the number of words in the \emph{target} summary and the number of words in the source. More compressive (lower ratio) data points are expected to be more challenging. 
    
\end{itemize}

\noindent Results are shown in \Cref{fig:3}, with one subplot for each of the four features and for each dataset. There are several important takeaways from this figure:

\begin{itemize}[leftmargin=*,topsep=2pt,itemsep=2pt,parsep=0pt]
    \item \textbf{SummaFusion is indeed better on lower quality base candidates} (top left subfigures). On every dataset, the green curve is significantly ahead of the blue one for summary bins of lowest quality. In fact, we notice that SummaFusion is even \emph{harmful} for summary bins of the highest quality (top 20\%). 
    
    \item \textbf{SummaFusion is better on more diverse base candidates} (top right subfigures). This is true as both the base model and SummaFusion perform worse when diversity increases, yet SummaFusion cushions the drop.
    
    \item \textbf{SummaFusion is better on longer source documents} (bottom left subfigures). This observation and the precedent confirm the hypothesis that SummaFusion helps on more challenging setups. 
    
    \item \textbf{SummaFusion is better on \emph{longer} summaries (on Reddit-TIFU)} (bottom right subfigures). There is no clear trend over all datasets for this feature. It is also not intuitive which case is harder to learn, as a short compression ratio means a higher level of summarizing, while a longer one corresponds to longer output summaries, which is also more prone to decoding errors. 
    
\end{itemize}

\begin{table}[]
\setlength{\tabcolsep}{0.2em}
\resizebox{\columnwidth}{!}{
\begin{tabular}{llcccccc}
\toprule
&                                  
& \multicolumn{3}{c}{\textbf{\emph{Source-abstractiveness}}}                                           
& \multicolumn{3}{c}{\textbf{\emph{Candidates-abstractiveness}}} \\
\multirow{-2}{*}{\textbf{Dataset}} 
& \multirow{-2}{*}{\textbf{Model}} 
& \textbf{1-grams} & \textbf{2-grams} & \textbf{3-grams} & \textbf{1-grams} & \textbf{2-grams} & \textbf{3-grams} \\

\midrule

\multirow{4}{*}{\textbf{XSum}} & Ground truth & 33.79 & 83.29 & 95.51 & \_ & \_ & \_ \\
                                & PEGASUS & 27.38 & 76.79 & 91.53 & \_ & \_ & \_ \\
                                & PEGASUS - \emph{oracle}  & \emph{28.53} & \emph{78.52} & \emph{93.06} & \_ & \_  & \_ \\
                                & SummaFusion-large & 27.18 & 75.71 & 90.94 & 5.46 & 16.62 & 25.78 \\

\midrule

\multirow{4}{*}{\textbf{Reddit-TIFU}} & Ground truth & 28.77 & 77.43 & 92.48 & \_ & \_ & \_ \\
                                      & PEGASUS & 12.96 & 57.20 & 78.72 & \_ & \_  & \_ \\
                                      & PEGASUS - \emph{oracle}  & \emph{14.19} & \emph{60.92} & \emph{82.86} & \_ & \_  & \_ \\
                                      & SummaFusion-large & 10.26 & 48.85 & 69.62 & 21.23 & 46.31 & 59.88 \\

\midrule

\multirow{4}{*}{\textbf{SAMSum}}      & Ground truth & 34.13 & 79.31 & 90.51 & \_ & \_ & \_ \\
                                      & PEGASUS & 25.06 & 68.99 & 82.41 & \_ & \_  & \_  \\
                                      & PEGASUS - \emph{oracle}  & \emph{26.88} & \emph{71.70} & \emph{85.08} & \_ & \_  & \_  \\
                                      & SummaFusion-large & 23.68 & 65.90	& 79.78 & 4.28 & 13.66 & 21.91 \\

\bottomrule
\end{tabular}
}
\caption{\textbf{Abstractiveness}. We report proportions of novel n-grams on the test set of each dataset, with regards to both the source and the sets of candidates.}
\label{tab:5}
\end{table}

\begin{table}[t]
\resizebox{\columnwidth}{!}{
\begin{tabular}{lccccc}
\toprule
\textbf{Model}              
& \textbf{R-1}
& \textbf{R-2}
& \textbf{R-L}
& \textbf{\begin{tabular}[c]{@{}c@{}}New \emph{source} \\ 2-grams\end{tabular}} 
& \textbf{\begin{tabular}[c]{@{}c@{}}New \emph{candidates} \\ 2-grams\end{tabular}} \\
\midrule
SummaFusion-large            & \textbf{30.08} & \textbf{10.48} & \textbf{23.99} & 48.85 & 46.31 \\
- candidates classification  & 29.27 & 10.24 & 23.47 & 48.60 & 41.45 \\
- input dropout              & 29.26 & 10.20 & 23.39 & 46.71 & 45.95 \\
- position token             & 29.23 & 10.31 & 23.40 & 45.93 & 44.44 \\
\hdashline 
Concat-baseline              & 26.87 & 8.48 & 21.59 & 64.78 & 13.27  \\
PEGASUS                      & 25.67 & 8.07 & 20.97 & 57.20 & \_ \\
\bottomrule
\end{tabular}
}
\caption{\textbf{Model ablation} study on Reddit-TIFU. We cumulatively remove components of SummaFusion, and report results on the test set.}
\label{tab:6}
\end{table}

\subsection{Abstractiveness}

Because SummaFusion conditions on both the source and the first-stage candidates, we shall now distinguish between two types of abstractiveness:

\begin{itemize}[leftmargin=*,topsep=2pt,itemsep=2pt,parsep=0pt]
    \item \emph{Source-abstractiveness}: this is the fraction of novel n-grams in the SummaFusion with regards to the \emph{source document}.
    
    \item \emph{Candidates-abstractiveness}: this is the fraction of novel n-grams in the SummaFusion with regards to the \emph{entire pool of candidates}.
\end{itemize}

We analyze both abstractiveness in \Cref{tab:5}, comparing with the base PEGASUS, and also the ground truth summaries (for \emph{source-abstractiveness}). Following other work, we measure abstractiveness with 1/2/3-grams counts. Surprisingly, SummaFusion is \emph{not} more abstractive with regards to the source, as it is slightly less abstractive than PEGASUS. Rather, SummaFusion is maintaining a high level of \emph{source-abstractiveness} on these highly abstractive datasets, and also offers a satisfactory level of \emph{candidates-abstractiveness}.

\begin{table}[]
\setlength{\tabcolsep}{0.2em}
\resizebox{\columnwidth}{!}{
\begin{tabular}{llccc}
\toprule
\textbf{Dataset} & \textbf{Model} & \textbf{R-1} & \textbf{R-2} & \textbf{R-L} \\

\midrule

\multirow{3}{*}{\textbf{XSum}} & SummaFusion & \textbf{47.08} & \textbf{24.05} & \textbf{38.82} \\ 
                               & SummaFusion - \emph{no source} & 46.64 & 23.55 & 38.30 \\
                               & SummaFusion -\emph{ no candidate} & 42.33 & 19.27 & 34.06 \\

\midrule

\multirow{3}{*}{\textbf{Reddit-TIFU}} & SummaFusion & \textbf{30.08} & \textbf{10.48} & \textbf{23.99} \\ 
                               & SummaFusion - \emph{no source} & 26.92 & 8.43 & 21.77 \\
                               & SummaFusion - \emph{no candidate} & 29.47 & 10.08 & 23.46 \\

\midrule

\multirow{3}{*}{\textbf{SAMSum}} & SummaFusion & \textbf{52.76} & \textbf{28.24} &  \textbf{43.98}  \\ 
                               & SummaFusion - \emph{no source} & 50.95 & 25.87 & 42.04 \\
                               & SummaFusion - \emph{no candidate} & 52.72 & 28.03 & 43.87 \\

\bottomrule
\end{tabular}
}
\caption{\textbf{Input ablation} on all datasets. We experiment with removing either the source or the entire set of candidates at inference.}
\label{tab:7}
\end{table}


\begin{figure}
    \centering 
    
    \minipage{0.33\columnwidth}
      \includegraphics[width=\linewidth]{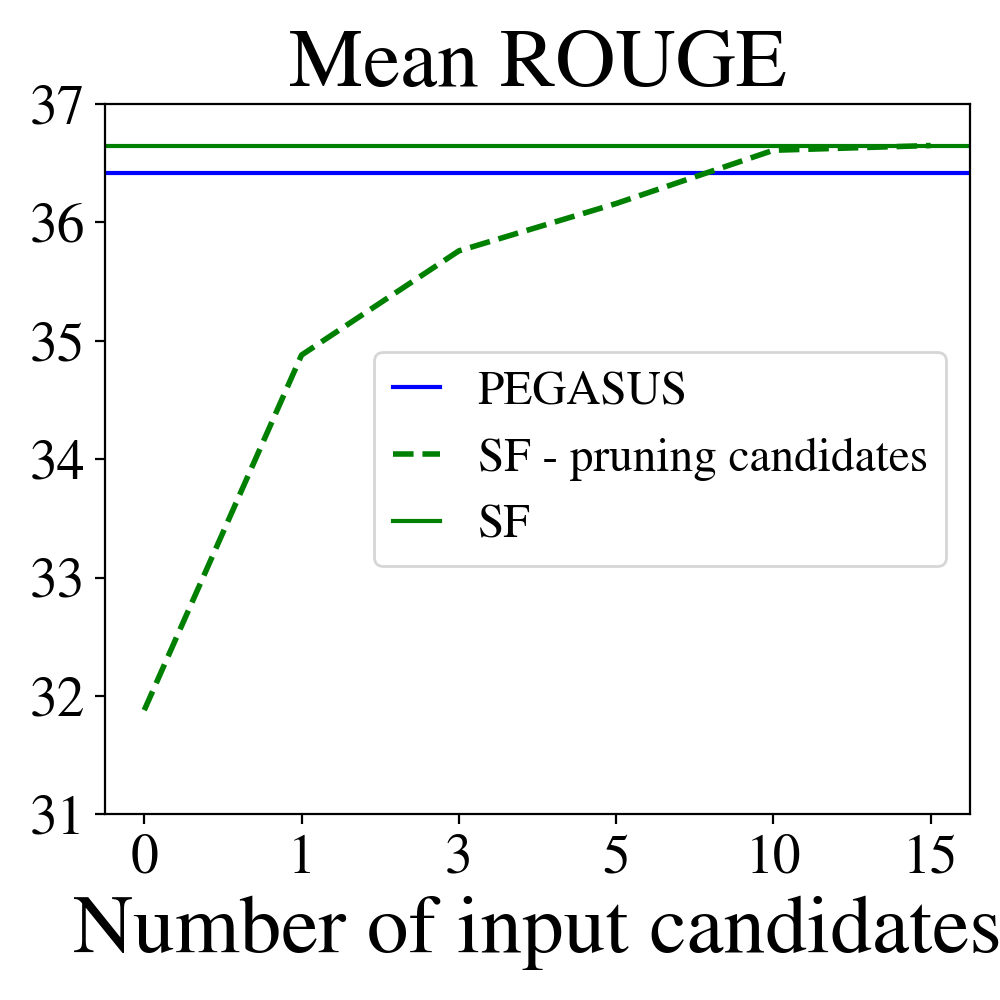}
      \caption*{XSum}
      \label{fig:4_1}
    \endminipage\hfill
    \minipage{0.33\columnwidth}
      \includegraphics[width=\linewidth]{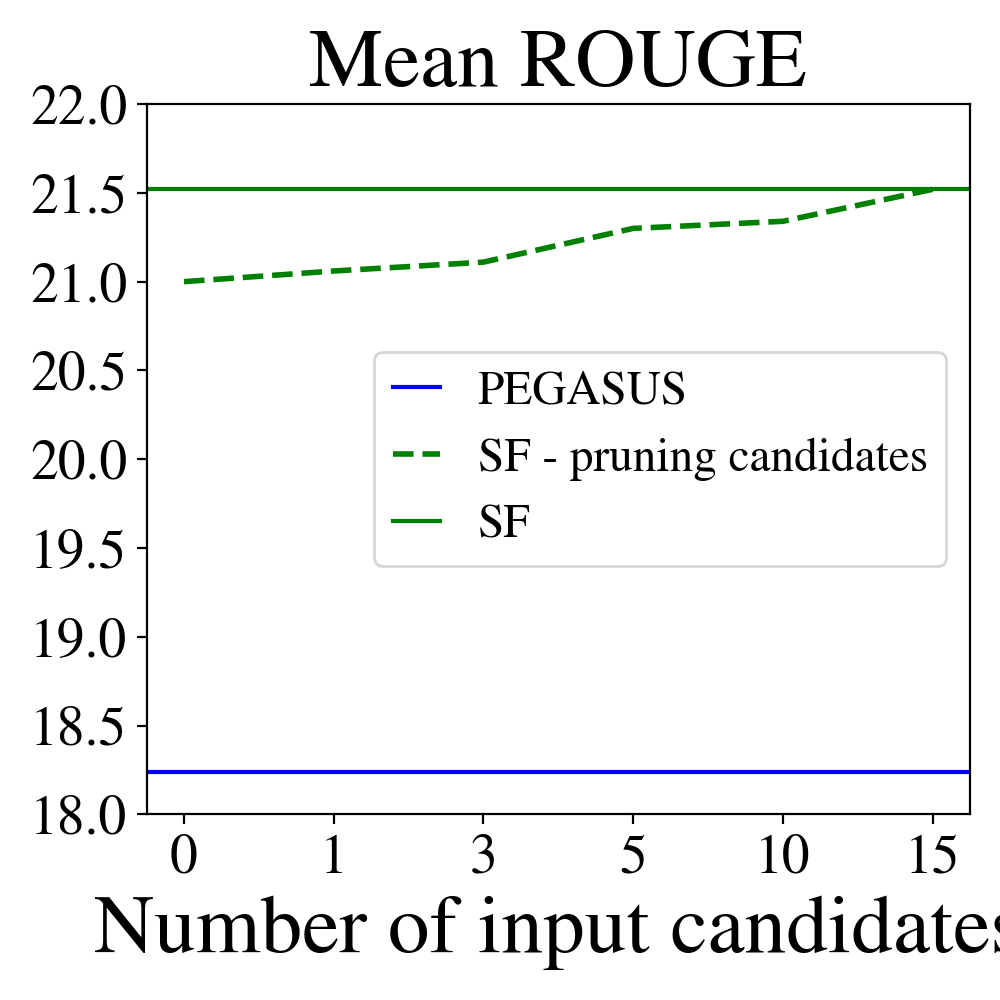}
      \caption*{Reddit-TIFU}
      \label{fig:4_2}
    \endminipage\hfill
    \minipage{0.33\columnwidth}%
      \includegraphics[width=\linewidth]{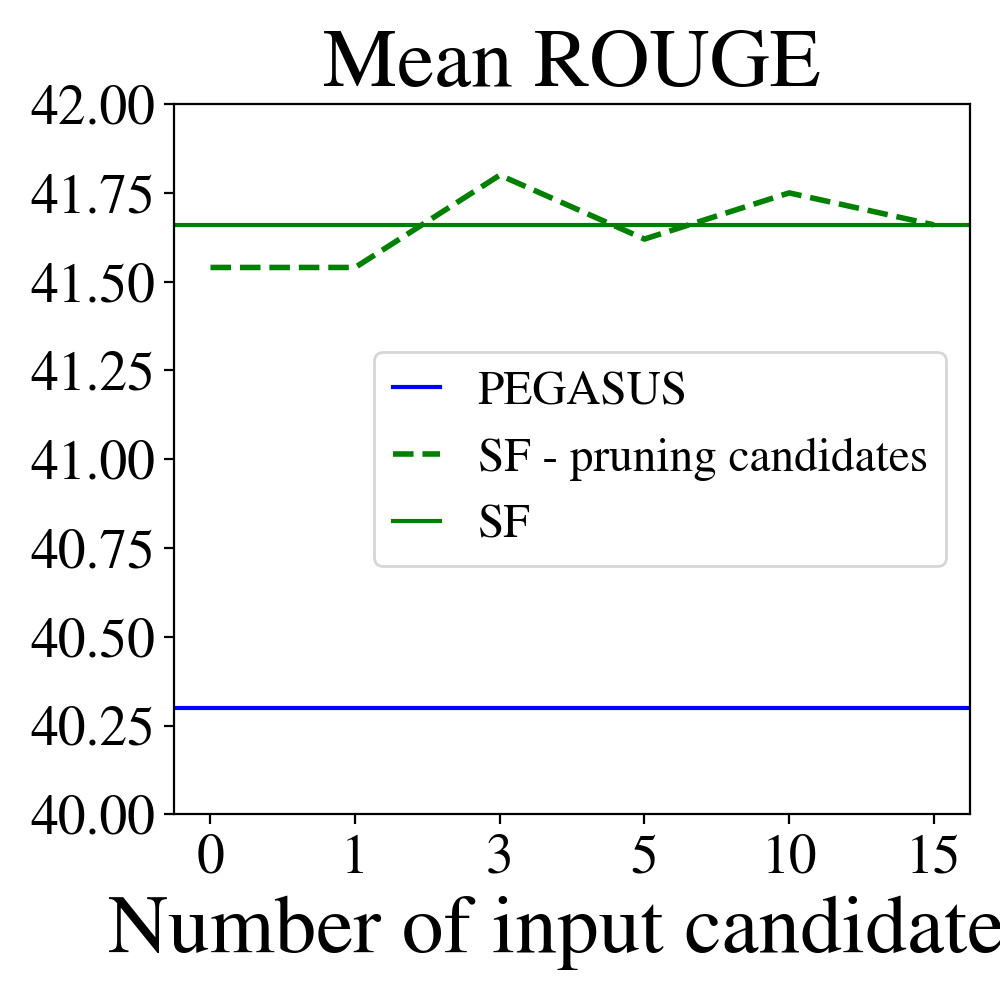}
      \caption*{SAMSum}
      \label{fig:4_3}
    \endminipage
    \caption{\textbf{Pruning input candidates} on the three datasets. We make inference with SummaFusion with a gradually increasing number of first-stage summary candidates. \textbf{SF} is SummaFusion.}
    \label{fig:4}
    \vspace{-1.0em}
\end{figure}

\subsection{Ablation}

To better understand how components of the model are interacting with each other in SummaFusion, we run an ablation study. We also compare our model with a naive baseline simply concatenating the truncated source and all summary candidates, referred to as \emph{Concat-baseline}. Results are shown in \Cref{tab:6}. Compared to its ablated versions and the \emph{Concat-baseline}, SummaFusion is able to achieve much higher {\sc{Rouge}}  while maintaining high \emph{source abstractiveness} and \emph{candidates abstractiveness}. The \emph{Concat-baseline}, despite reaching good \emph{source abstractiveness}, is not able to produce a satisfactory level of \emph{candidates abstractiveness} (only 13.27\% new 2-grams compared to 46.31\% for SummaFusion).

To assert the importance of each input stream, we perform inference when removing either the source, or the entire set of first-stage candidates. Removal is done by replacing inputs with the tokens used during input dropout \Cref{sec:3_3}. As we can see in \Cref{tab:7}, both the source and the candidates are highly necessary for SummaFusion to reach full performance. \Cref{fig:4} provides finer-grained insights into how performance varies with a gradually increasing number of input candidates. This confirms our choice of conditioning SummaFusion decoding on both the source and \emph{all} the first-stage candidates.

\begin{table}[t]
\resizebox{\columnwidth}{!}{
\begin{tabular}{lcccc}
\toprule
\multirow{3}{*}{\textbf{Summary}} 
& \multirow{3}{*}{\textbf{\begin{tabular}[c]{@{}c@{}}Overall\\ preference\end{tabular}}}
& \multicolumn{3}{c}{\textbf{Reasons}} \\
& & \begin{tabular}[c]{@{}c@{}}More\\ informative\end{tabular}
& \begin{tabular}[c]{@{}c@{}}More fluent \\ or grammatical\end{tabular}
& \begin{tabular}[c]{@{}c@{}}More factually\\ consistent\end{tabular} \\
\midrule
\multicolumn{5}{c}{\textit{\textbf{XSum}}} \\
\midrule

PEGASUS            & 15.33 (6.11) & 6.67 (2.52) & 3.00 (5.20) & 8.67 (3.79) \\
SummaFusion-large  & \textbf{24.33} (7.57) & \textbf{11.67} (2.31) & \textbf{8.33} (8.74) & \textbf{9.00} (2.65) \\
Tie                & 10.33 (8.08) & \_ & \_ & \_ \\
\hdashline 
SummaReranker      & \textbf{20.33} (3.21) & 11.00 (2.65) & 1.33 (1.53) & \textbf{9.67} (3.06) \\
SummaFusion-large  & \textbf{20.33} (0.58) & \textbf{13.00} (3.61) & \textbf{2.33} (2.31) & \textbf{9.67} (1.53) \\
Tie                & 9.33 (3.06) & \_ & \_ & \_ \\

\midrule
\multicolumn{5}{c}{\textit{\textbf{Reddit-TIFU}}} \\
\midrule

PEGASUS            & 9.67	(1.53) & 5.33 (0.71) & 1.00 (1.41) & 5.00 (0.71) \\
SummaFusion-large  & \textbf{30.67} (3.79) & \textbf{24.33} (6.36) & \textbf{6.00} (2.83) & \textbf{16.67} (10.61) \\
Tie                & 9.67 (4.93) & \_ & \_ & \_ \\
\hdashline 
SummaReranker      & 12.67 (1.53) & 5.00 (2.12) & 2.33 (1.41) & 7.00 (2.12) \\ 
SummaFusion-large  & \textbf{32.33} (1.53) & \textbf{22.33} (1.41) & \textbf{5.00} (0.00) & \textbf{16.67} (0.71) \\
Tie                & 5.00 (1.00) & \_ & \_ & \_ \\

\midrule
\multicolumn{5}{c}{\textit{\textbf{SAMSum}}} \\
\midrule

PEGASUS            & 14.67 (1.53) & 9.33 (1.15) & 0.33 (0.58) & 8.33 (2.52) \\
SummaFusion-large  & \textbf{26.00} (4.36) & \textbf{18.67} (1.53) & \textbf{3.33} (1.15) & \textbf{12.00} (10.82) \\
Tie                & 9.33 (2.89) & \_ & \_ & \_ \\    
\hdashline 
SummaReranker      & 17.00 (1.00) & 8.67 (1.15) & 1.33 (1.53) & 5.33 (4.51) \\ 
SummaFusion-large  & \textbf{24.33} (1.53) & \textbf{16.67} (4.51) & \textbf{2.33} (1.53) & \textbf{10.33} (2.08) \\ 
Tie                & 8.67 (2.08) & \_ & \_ & \_ \\

\bottomrule
\end{tabular}
}
\caption{\textbf{Human evaluation} on all datasets. We show mean counts over three humans rating 50 data points in each dataset, with standard deviation in parenthesis. For each dataset, the first block compares the PEGASUS summary with the SummaFusion one, while the second block compares SummaReranker with SummaFusion.}
\label{tab:8}
\vspace{-1.0em}
\end{table}

\subsection{Human Evaluation}

We run a human study comparing a baseline summary with the SummaFusion one on 50 random samples from each dataset. As baseline, we use both PEGASUS \cite{zhang2020pegasus}, and SummaReranker \cite{ravaut-etal-2022-summareranker}, in order to compare to another second-stage method. Human volunteers are graduate students with professional English proficiency, and we select three volunteers per dataset. Human graders have to decide which summary they prefer or if it is a tie. In the former, they also have to select at least one reason motivating the preference among the three following: the summary is more \emph{informative}, more \emph{fluent or grammatical}, or more \emph{factually consistent} with the source.

As we see in \Cref{tab:8}, humans clearly prefer SummaFusion summaries over PEGASUS ones on all dataset. The difference is striking in terms of fluency, and informativeness on Reddit-TIFU. On XSum, SummaReranker and SummaFusion summaries are deemed of equal quality, but SummaFusion is preferred on the two other datasets.

\subsection{Breaking the Oracle Barrier}

Due to being an abstractive second-stage summarization method, SummaFusion is not bounded by the quality of the first-stage candidates, including even the (ranking) oracle. We investigate data points where SummaFusion can indeed surpass the first-stage oracle. Since SummaFusion's relative performance gain is greater with less labels (shown in \Cref{sec:few-shot}), we vary the available supervision. At the same time, decreasing the number of candidates $m$ decreases the oracle score providing a small headroom (see \Cref{sec:appendix_a} \Cref{tab:a_1_3}). Since during training, SummaFusion sees between 2 and 15 candidates, we have the flexibility to input any number of candidates in this range at inference. We experiment with $m \in \{5, 10, 15\}.$

Results on Reddit TIFU are summarized in \Cref{tab:9}. Impressively, SummaFusion can outperform the oracle in more than 30\% cases with 5 candidates. This unveils yet a new interesting use case for SummaFusion: if computational budget is limited at inference and the beam width is capped to a lower value, it is also very beneficial to use SummaFusion to improve on the generated summaries.

\begin{table}[t]
\resizebox{\columnwidth}{!}{
\begin{tabular}{lcccc}
\toprule
\multirow{2}{*}{\textbf{\begin{tabular}[c]{@{}l@{}}Number of\\ candidates (m)\end{tabular}}} 
& \multicolumn{4}{c}{\textbf{Available supervision}} \\
& 10-shot & 100-shot & 1000-shot & all data    \\
\midrule
$m$ = 5      & 45.78\% & 30.51\% & 30.08\% & 37.04\% \\
$m$ = 10     & 27.33\% & 21.34\% & 21.44\% & 28.66\% \\
$m$ = 15     & 17.50\% & 17.57\% & 16.72\% & 24.87\% \\
\bottomrule
\end{tabular}
}
\caption{\textbf{SummaFusion surpassing the first-stage oracle} counts (as percentages) on Reddit-TIFU test set. We count these cases across two dimensions: number of first-stage candidates, and available supervision.}
\label{tab:9}
\vspace{-1.0em}
\end{table}

\section{Conclusion}
\label{sec:conclusion}

We introduced SummaFusion, the first method for abstractive second-stage summarization. Our model encodes the source document and each diverse beam search summary candidate individually, and fuses them in the decoder. It is designed for very abstractive summarization tasks, and works especially well on challenging data points such as longer source documents. We achieve state-of-the-art {\sc{Rouge}} results in 10-shot-and 100-shot summarization on XSum, Reddit-TIFU and SAMSum. Besides, fused summaries are favored by humans over first-stage PEGASUS candidates.

\section*{Limitations}
As a second-stage abstractive summarization model, a drawback of our approach is that it requires to train an additional model on top of the base summarization model. Besides, during training, because we split the training set in halves, we actually require to train two base summarization models. We also need to generate summary candidates for each data point of the training, validation and test sets, which is time consuming. For these reasons, SummaFusion presents some computational overhead. Nevertheless, training and inference fits into a single Nvidia RTX 6000 24GB GPU. 

We also observed that SummaFusion worked less well with beam search candidates than diverse beam search ones. While we attribute this to beam search candidates being too similar with each other, it remains an open question how to improve SummaFusion on such cases with very similar input candidates.

\section*{Ethics Statement}

Our proposed approach is an abstractive summarization method. Therefore, it is prone to hallucinations and generating summaries with facts not in the source document, with some of these facts being wrong. Therefore, the model outputs should be analyzed with caution in critical scenarios.

 \section*{Acknowledgements}
This research was supported by the SINGA scholarship and partially supported by the National Re-
search Foundation, Prime Minister’s Office, Singapore under its Campus for Research Excellence and
Technological Enterprise (CREATE) programme. We would like to thank anonymous reviewers for
several very insightful feedback on how to improve the paper, especially with regards to ablations and comparisons to other second-stage summarization works. We thank Florian Le Bronnec for helpful proof-reading of the paper.

\bibliography{anthology,custom}

\begin{thebibliography}{38}
\expandafter\ifx\csname natexlab\endcsname\relax\def\natexlab#1{#1}\fi

\bibitem[{Barzilay and McKeown(2005)}]{barzilay-mckeown-2005-sentence}
Regina Barzilay and Kathleen~R. McKeown. 2005.
\newblock \href {https://doi.org/10.1162/089120105774321091} {Sentence fusion
  for multidocument news summarization}.
\newblock \emph{Computational Linguistics}, 31(3):297--328.

\bibitem[{Bengio et~al.(2015)Bengio, Vinyals, Jaitly, and
  Shazeer}]{Bengio-NIPS-15}
Samy Bengio, Oriol Vinyals, Navdeep Jaitly, and Noam Shazeer. 2015.
\newblock Scheduled sampling for sequence prediction with recurrent neural
  networks.
\newblock In \emph{Proceedings of the 28th International Conference on Neural
  Information Processing Systems - Volume 1}, NIPS'15, page 1171–1179,
  Cambridge, MA, USA. MIT Press.

\bibitem[{Cao et~al.(2018)Cao, Li, Li, and Wei}]{cao-etal-2018-retrieve}
Ziqiang Cao, Wenjie Li, Sujian Li, and Furu Wei. 2018.
\newblock \href {https://doi.org/10.18653/v1/P18-1015} {Retrieve, rerank and
  rewrite: Soft template based neural summarization}.
\newblock In \emph{Proceedings of the 56th Annual Meeting of the Association
  for Computational Linguistics (Volume 1: Long Papers)}, pages 152--161,
  Melbourne, Australia. Association for Computational Linguistics.

\bibitem[{Dou et~al.(2021)Dou, Liu, Hayashi, Jiang, and
  Neubig}]{dou-etal-2021-gsum}
Zi-Yi Dou, Pengfei Liu, Hiroaki Hayashi, Zhengbao Jiang, and Graham Neubig.
  2021.
\newblock \href {https://doi.org/10.18653/v1/2021.naacl-main.384} {{GS}um: A
  general framework for guided neural abstractive summarization}.
\newblock In \emph{Proceedings of the 2021 Conference of the North American
  Chapter of the Association for Computational Linguistics: Human Language
  Technologies}, pages 4830--4842, Online. Association for Computational
  Linguistics.

\bibitem[{Fabbri et~al.(2021)Fabbri, Han, Li, Li, Ghazvininejad, Joty, Radev,
  and Mehdad}]{fabbri-etal-2021-improving}
Alexander Fabbri, Simeng Han, Haoyuan Li, Haoran Li, Marjan Ghazvininejad,
  Shafiq Joty, Dragomir Radev, and Yashar Mehdad. 2021.
\newblock \href {https://doi.org/10.18653/v1/2021.naacl-main.57} {Improving
  zero and few-shot abstractive summarization with intermediate fine-tuning and
  data augmentation}.
\newblock In \emph{Proceedings of the 2021 Conference of the North American
  Chapter of the Association for Computational Linguistics: Human Language
  Technologies}, pages 704--717, Online. Association for Computational
  Linguistics.

\bibitem[{Gliwa et~al.(2019)Gliwa, Mochol, Biesek, and
  Wawer}]{gliwa-etal-2019-samsum}
Bogdan Gliwa, Iwona Mochol, Maciej Biesek, and Aleksander Wawer. 2019.
\newblock \href {https://doi.org/10.18653/v1/D19-5409} {{SAMS}um corpus: A
  human-annotated dialogue dataset for abstractive summarization}.
\newblock In \emph{Proceedings of the 2nd Workshop on New Frontiers in
  Summarization}, pages 70--79, Hong Kong, China. Association for Computational
  Linguistics.

\bibitem[{Hermann et~al.(2015)Hermann, Kocisky, Grefenstette, Espeholt, Kay,
  Suleyman, and Blunsom}]{hermann2015teaching}
Karl~Moritz Hermann, Tomas Kocisky, Edward Grefenstette, Lasse Espeholt, Will
  Kay, Mustafa Suleyman, and Phil Blunsom. 2015.
\newblock Teaching machines to read and comprehend.
\newblock \emph{Advances in neural information processing systems},
  28:1693--1701.

\bibitem[{Holtzman et~al.(2019)Holtzman, Buys, Du, Forbes, and
  Choi}]{holtzman2019curious}
Ari Holtzman, Jan Buys, Li~Du, Maxwell Forbes, and Yejin Choi. 2019.
\newblock The curious case of neural text degeneration.
\newblock \emph{arXiv preprint arXiv:1904.09751}.

\bibitem[{Izacard and Grave(2021)}]{izacard-grave-2021-leveraging}
Gautier Izacard and Edouard Grave. 2021.
\newblock \href {https://doi.org/10.18653/v1/2021.eacl-main.74} {Leveraging
  passage retrieval with generative models for open domain question answering}.
\newblock In \emph{Proceedings of the 16th Conference of the European Chapter
  of the Association for Computational Linguistics: Main Volume}, pages
  874--880, Online. Association for Computational Linguistics.

\bibitem[{Kim et~al.(2019)Kim, Kim, and Kim}]{kim-etal-2019-abstractive}
Byeongchang Kim, Hyunwoo Kim, and Gunhee Kim. 2019.
\newblock \href {https://doi.org/10.18653/v1/N19-1260} {Abstractive
  summarization of {R}eddit posts with multi-level memory networks}.
\newblock In \emph{Proceedings of the 2019 Conference of the North {A}merican
  Chapter of the Association for Computational Linguistics: Human Language
  Technologies, Volume 1 (Long and Short Papers)}, pages 2519--2531,
  Minneapolis, Minnesota. Association for Computational Linguistics.

\bibitem[{Kingma and Ba(2014)}]{kingma2014adam}
Diederik~P Kingma and Jimmy Ba. 2014.
\newblock Adam: A method for stochastic optimization.
\newblock \emph{arXiv preprint arXiv:1412.6980}.

\bibitem[{Lebanoff et~al.(2020{\natexlab{a}})Lebanoff, Dernoncourt, Kim, Chang,
  and Liu}]{lebanoff-etal-2020-cascade}
Logan Lebanoff, Franck Dernoncourt, Doo~Soon Kim, Walter Chang, and Fei Liu.
  2020{\natexlab{a}}.
\newblock \href {https://aclanthology.org/2020.aacl-main.52} {A cascade
  approach to neural abstractive summarization with content selection and
  fusion}.
\newblock In \emph{Proceedings of the 1st Conference of the Asia-Pacific
  Chapter of the Association for Computational Linguistics and the 10th
  International Joint Conference on Natural Language Processing}, pages
  529--535, Suzhou, China. Association for Computational Linguistics.

\bibitem[{Lebanoff et~al.(2020{\natexlab{b}})Lebanoff, Dernoncourt, Kim, Wang,
  Chang, and Liu}]{lebanoff-etal-2020-learning}
Logan Lebanoff, Franck Dernoncourt, Doo~Soon Kim, Lidan Wang, Walter Chang, and
  Fei Liu. 2020{\natexlab{b}}.
\newblock \href {https://doi.org/10.18653/v1/2020.emnlp-main.338} {Learning to
  fuse sentences with transformers for summarization}.
\newblock In \emph{Proceedings of the 2020 Conference on Empirical Methods in
  Natural Language Processing (EMNLP)}, pages 4136--4142, Online. Association
  for Computational Linguistics.

\bibitem[{Lebanoff et~al.(2019)Lebanoff, Muchovej, Dernoncourt, Kim, Kim,
  Chang, and Liu}]{lebanoff-etal-2019-analyzing}
Logan Lebanoff, John Muchovej, Franck Dernoncourt, Doo~Soon Kim, Seokhwan Kim,
  Walter Chang, and Fei Liu. 2019.
\newblock \href {https://doi.org/10.18653/v1/D19-5413} {Analyzing sentence
  fusion in abstractive summarization}.
\newblock In \emph{Proceedings of the 2nd Workshop on New Frontiers in
  Summarization}, pages 104--110, Hong Kong, China. Association for
  Computational Linguistics.

\bibitem[{Lewis et~al.(2020)Lewis, Liu, Goyal, Ghazvininejad, Mohamed, Levy,
  Stoyanov, and Zettlemoyer}]{lewis-etal-2020-bart}
Mike Lewis, Yinhan Liu, Naman Goyal, Marjan Ghazvininejad, Abdelrahman Mohamed,
  Omer Levy, Veselin Stoyanov, and Luke Zettlemoyer. 2020.
\newblock \href {https://doi.org/10.18653/v1/2020.acl-main.703} {{BART}:
  Denoising sequence-to-sequence pre-training for natural language generation,
  translation, and comprehension}.
\newblock In \emph{Proceedings of the 58th Annual Meeting of the Association
  for Computational Linguistics}, pages 7871--7880, Online. Association for
  Computational Linguistics.

\bibitem[{Lhoest et~al.(2021)Lhoest, Villanova~del Moral, Jernite, Thakur, von
  Platen, Patil, Chaumond, Drame, Plu, Tunstall, Davison, {\v{S}}a{\v{s}}ko,
  Chhablani, Malik, Brandeis, Le~Scao, Sanh, Xu, Patry, McMillan-Major, Schmid,
  Gugger, Delangue, Matussi{\`e}re, Debut, Bekman, Cistac, Goehringer, Mustar,
  Lagunas, Rush, and Wolf}]{lhoest-etal-2021-datasets}
Quentin Lhoest, Albert Villanova~del Moral, Yacine Jernite, Abhishek Thakur,
  Patrick von Platen, Suraj Patil, Julien Chaumond, Mariama Drame, Julien Plu,
  Lewis Tunstall, Joe Davison, Mario {\v{S}}a{\v{s}}ko, Gunjan Chhablani,
  Bhavitvya Malik, Simon Brandeis, Teven Le~Scao, Victor Sanh, Canwen Xu,
  Nicolas Patry, Angelina McMillan-Major, Philipp Schmid, Sylvain Gugger,
  Cl{\'e}ment Delangue, Th{\'e}o Matussi{\`e}re, Lysandre Debut, Stas Bekman,
  Pierric Cistac, Thibault Goehringer, Victor Mustar, Fran{\c{c}}ois Lagunas,
  Alexander Rush, and Thomas Wolf. 2021.
\newblock \href {https://doi.org/10.18653/v1/2021.emnlp-demo.21} {Datasets: A
  community library for natural language processing}.
\newblock In \emph{Proceedings of the 2021 Conference on Empirical Methods in
  Natural Language Processing: System Demonstrations}, pages 175--184, Online
  and Punta Cana, Dominican Republic. Association for Computational
  Linguistics.

\bibitem[{Lin(2004)}]{lin-2004-rouge}
Chin-Yew Lin. 2004.
\newblock \href {https://aclanthology.org/W04-1013} {{ROUGE}: A package for
  automatic evaluation of summaries}.
\newblock In \emph{Text Summarization Branches Out}, pages 74--81, Barcelona,
  Spain. Association for Computational Linguistics.

\bibitem[{Liu et~al.(2022{\natexlab{a}})Liu, Bai, Li, Hu, and Gao}]{liu2022psp}
Xiaochen Liu, Yu~Bai, Jiawei Li, Yinan Hu, and Yang Gao. 2022{\natexlab{a}}.
\newblock Psp: Pre-trained soft prompts for few-shot abstractive summarization.
\newblock \emph{arXiv preprint arXiv:2204.04413}.

\bibitem[{Liu et~al.(2021)Liu, Dou, and Liu}]{liu-etal-2021-refsum}
Yixin Liu, Zi-Yi Dou, and Pengfei Liu. 2021.
\newblock \href {https://doi.org/10.18653/v1/2021.naacl-main.113} {{R}ef{S}um:
  Refactoring neural summarization}.
\newblock In \emph{Proceedings of the 2021 Conference of the North American
  Chapter of the Association for Computational Linguistics: Human Language
  Technologies}, pages 1437--1448, Online. Association for Computational
  Linguistics.

\bibitem[{Liu and Liu(2021)}]{liu-liu-2021-simcls}
Yixin Liu and Pengfei Liu. 2021.
\newblock \href {https://doi.org/10.18653/v1/2021.acl-short.135} {{S}im{CLS}: A
  simple framework for contrastive learning of abstractive summarization}.
\newblock In \emph{Proceedings of the 59th Annual Meeting of the Association
  for Computational Linguistics and the 11th International Joint Conference on
  Natural Language Processing (Volume 2: Short Papers)}, pages 1065--1072,
  Online. Association for Computational Linguistics.

\bibitem[{Liu et~al.(2022{\natexlab{b}})Liu, Liu, Radev, and
  Neubig}]{liu-etal-2022-brio}
Yixin Liu, Pengfei Liu, Dragomir Radev, and Graham Neubig. 2022{\natexlab{b}}.
\newblock \href {https://doi.org/10.18653/v1/2022.acl-long.207} {{BRIO}:
  Bringing order to abstractive summarization}.
\newblock In \emph{Proceedings of the 60th Annual Meeting of the Association
  for Computational Linguistics (Volume 1: Long Papers)}, pages 2890--2903,
  Dublin, Ireland. Association for Computational Linguistics.

\bibitem[{Narayan et~al.(2018)Narayan, Cohen, and
  Lapata}]{narayan-etal-2018-dont}
Shashi Narayan, Shay~B. Cohen, and Mirella Lapata. 2018.
\newblock \href {https://doi.org/10.18653/v1/D18-1206} {Don{'}t give me the
  details, just the summary! topic-aware convolutional neural networks for
  extreme summarization}.
\newblock In \emph{Proceedings of the 2018 Conference on Empirical Methods in
  Natural Language Processing}, pages 1797--1807, Brussels, Belgium.
  Association for Computational Linguistics.

\bibitem[{Narayan et~al.(2022)Narayan, Sim{\~o}es, Zhao, Maynez, Das, Collins,
  and Lapata}]{narayan-etal-2022-well}
Shashi Narayan, Gon{\c{c}}alo Sim{\~o}es, Yao Zhao, Joshua Maynez, Dipanjan
  Das, Michael Collins, and Mirella Lapata. 2022.
\newblock \href {https://doi.org/10.18653/v1/2022.acl-long.94} {A well-composed
  text is half done! composition sampling for diverse conditional generation}.
\newblock In \emph{Proceedings of the 60th Annual Meeting of the Association
  for Computational Linguistics (Volume 1: Long Papers)}, pages 1319--1339,
  Dublin, Ireland. Association for Computational Linguistics.

\bibitem[{Provilkov et~al.(2020)Provilkov, Emelianenko, and
  Voita}]{provilkov-etal-2020-bpe}
Ivan Provilkov, Dmitrii Emelianenko, and Elena Voita. 2020.
\newblock \href {https://doi.org/10.18653/v1/2020.acl-main.170} {{BPE}-dropout:
  Simple and effective subword regularization}.
\newblock In \emph{Proceedings of the 58th Annual Meeting of the Association
  for Computational Linguistics}, pages 1882--1892, Online. Association for
  Computational Linguistics.

\bibitem[{Qi et~al.(2021)Qi, Gong, Yan, Xu, Yao, Zhou, Cheng, Jiang, Chen,
  Zhang, Li, and Duan}]{qi-etal-2021-prophetnet}
Weizhen Qi, Yeyun Gong, Yu~Yan, Can Xu, Bolun Yao, Bartuer Zhou, Biao Cheng,
  Daxin Jiang, Jiusheng Chen, Ruofei Zhang, Houqiang Li, and Nan Duan. 2021.
\newblock \href {https://doi.org/10.18653/v1/2021.acl-demo.28}
  {{P}rophet{N}et-{X}: Large-scale pre-training models for {E}nglish,
  {C}hinese, multi-lingual, dialog, and code generation}.
\newblock In \emph{Proceedings of the 59th Annual Meeting of the Association
  for Computational Linguistics and the 11th International Joint Conference on
  Natural Language Processing: System Demonstrations}, pages 232--239, Online.
  Association for Computational Linguistics.

\bibitem[{Raffel et~al.(2019)Raffel, Shazeer, Roberts, Lee, Narang, Matena,
  Zhou, Li, and Liu}]{raffel2019exploring}
Colin Raffel, Noam Shazeer, Adam Roberts, Katherine Lee, Sharan Narang, Michael
  Matena, Yanqi Zhou, Wei Li, and Peter~J Liu. 2019.
\newblock Exploring the limits of transfer learning with a unified text-to-text
  transformer.
\newblock \emph{arXiv preprint arXiv:1910.10683}.

\bibitem[{Ravaut et~al.(2022)Ravaut, Joty, and
  Chen}]{ravaut-etal-2022-summareranker}
Mathieu Ravaut, Shafiq Joty, and Nancy Chen. 2022.
\newblock \href {https://doi.org/10.18653/v1/2022.acl-long.309}
  {{S}umma{R}eranker: A multi-task mixture-of-experts re-ranking framework for
  abstractive summarization}.
\newblock In \emph{Proceedings of the 60th Annual Meeting of the Association
  for Computational Linguistics (Volume 1: Long Papers)}, pages 4504--4524,
  Dublin, Ireland. Association for Computational Linguistics.

\bibitem[{Reddy(1977)}]{reddy1977speech}
Raj Reddy. 1977.
\newblock Speech understanding systems: A summary of results of the five-year
  research effort at carnegie mellon university.
\newblock \emph{Pittsburgh, Pa}.

\bibitem[{See et~al.(2017)See, Liu, and Manning}]{see-etal-2017-get}
Abigail See, Peter~J. Liu, and Christopher~D. Manning. 2017.
\newblock \href {https://doi.org/10.18653/v1/P17-1099} {Get to the point:
  Summarization with pointer-generator networks}.
\newblock In \emph{Proceedings of the 55th Annual Meeting of the Association
  for Computational Linguistics (Volume 1: Long Papers)}, pages 1073--1083,
  Vancouver, Canada. Association for Computational Linguistics.

\bibitem[{Sun and Li(2021)}]{sun2021alleviating}
Shichao Sun and Wenjie Li. 2021.
\newblock Alleviating exposure bias via contrastive learning for abstractive
  text summarization.
\newblock \emph{arXiv preprint arXiv:2108.11846}.

\bibitem[{Vaswani et~al.(2017)Vaswani, Shazeer, Parmar, Uszkoreit, Jones,
  Gomez, Kaiser, and Polosukhin}]{vaswani2017attention}
Ashish Vaswani, Noam Shazeer, Niki Parmar, Jakob Uszkoreit, Llion Jones,
  Aidan~N Gomez, {\L}ukasz Kaiser, and Illia Polosukhin. 2017.
\newblock Attention is all you need.
\newblock \emph{Advances in neural information processing systems}, 30.

\bibitem[{Vijayakumar et~al.(2016)Vijayakumar, Cogswell, Selvaraju, Sun, Lee,
  Crandall, and Batra}]{vijayakumar2016diverse}
Ashwin~K Vijayakumar, Michael Cogswell, Ramprasath~R Selvaraju, Qing Sun,
  Stefan Lee, David Crandall, and Dhruv Batra. 2016.
\newblock Diverse beam search: Decoding diverse solutions from neural sequence
  models.
\newblock \emph{arXiv preprint arXiv:1610.02424}.

\bibitem[{Weiss et~al.(2021)Weiss, Roit, Ernst, and Dagan}]{weiss2021extending}
Daniela~Brook Weiss, Paul Roit, Ori Ernst, and Ido Dagan. 2021.
\newblock Extending multi-text sentence fusion resources via pyramid
  annotations.
\newblock \emph{arXiv preprint arXiv:2110.04517}.

\bibitem[{Williams and Zipser(1989)}]{williams1989learning}
Ronald~J Williams and David Zipser. 1989.
\newblock A learning algorithm for continually running fully recurrent neural
  networks.
\newblock \emph{Neural computation}, 1(2):270--280.

\bibitem[{Wolf et~al.(2020)Wolf, Debut, Sanh, Chaumond, Delangue, Moi, Cistac,
  Rault, Louf, Funtowicz, Davison, Shleifer, von Platen, Ma, Jernite, Plu, Xu,
  Le~Scao, Gugger, Drame, Lhoest, and Rush}]{wolf-etal-2020-transformers}
Thomas Wolf, Lysandre Debut, Victor Sanh, Julien Chaumond, Clement Delangue,
  Anthony Moi, Pierric Cistac, Tim Rault, Remi Louf, Morgan Funtowicz, Joe
  Davison, Sam Shleifer, Patrick von Platen, Clara Ma, Yacine Jernite, Julien
  Plu, Canwen Xu, Teven Le~Scao, Sylvain Gugger, Mariama Drame, Quentin Lhoest,
  and Alexander Rush. 2020.
\newblock \href {https://doi.org/10.18653/v1/2020.emnlp-demos.6} {Transformers:
  State-of-the-art natural language processing}.
\newblock In \emph{Proceedings of the 2020 Conference on Empirical Methods in
  Natural Language Processing: System Demonstrations}, pages 38--45, Online.
  Association for Computational Linguistics.

\bibitem[{Xu et~al.(2021)Xu, Zhang, Wu, and Wei}]{xu2021sequence}
Shusheng Xu, Xingxing Zhang, Yi~Wu, and Furu Wei. 2021.
\newblock Sequence level contrastive learning for text summarization.
\newblock \emph{arXiv preprint arXiv:2109.03481}.

\bibitem[{Yang et~al.(2020)Yang, Zhu, Gmyr, Zeng, Huang, and
  Darve}]{yang-etal-2020-ted}
Ziyi Yang, Chenguang Zhu, Robert Gmyr, Michael Zeng, Xuedong Huang, and Eric
  Darve. 2020.
\newblock \href {https://doi.org/10.18653/v1/2020.findings-emnlp.168} {{TED}: A
  pretrained unsupervised summarization model with theme modeling and
  denoising}.
\newblock In \emph{Findings of the Association for Computational Linguistics:
  EMNLP 2020}, pages 1865--1874, Online. Association for Computational
  Linguistics.

\bibitem[{Zhang et~al.(2020)Zhang, Zhao, Saleh, and Liu}]{zhang2020pegasus}
Jingqing Zhang, Yao Zhao, Mohammad Saleh, and Peter Liu. 2020.
\newblock Pegasus: Pre-training with extracted gap-sentences for abstractive
  summarization.
\newblock In \emph{International Conference on Machine Learning}, pages
  11328--11339. PMLR.

\end{thebibliography}
\bibliographystyle{acl_natbib}

\appendix

\newpage
\section{Oracle Scores}
\label{sec:appendix_a}

We show oracle scores (in terms of {\sc{Rouge-1/2/L}}) results for the same PEGASUS decoded with 15 beams for both beam search and diverse beam search.  

\begin{table}[h]
\resizebox{\columnwidth}{!}{
\begin{tabular}{llccc}
\toprule
\textbf{Dataset}             
& \textbf{\begin{tabular}[c]{@{}l@{}}Decoding\\ method\end{tabular}} 
& \textbf{\begin{tabular}[c]{@{}c@{}}R-1\\ oracle\end{tabular}} 
& \textbf{\begin{tabular}[c]{@{}c@{}}R-2\\ oracle\end{tabular}} 
& \textbf{\begin{tabular}[c]{@{}c@{}}R-L\\ oracle\end{tabular}} \\
\midrule
\multirow{2}{*}{XSum}        & Beam search              & 56.07 & 33.80 & 48.33 \\
                             & Diverse beam search      & 57.82 & 35.28 & 50.75 \\
\midrule
\multirow{2}{*}{Reddit-TIFU} & Beam search              & 36.08 & 14.93 & 29.70 \\
                             & Diverse beam search      & 36.70 & 15.22 & 30.88 \\
\midrule
\multirow{2}{*}{SAMSum}      & Beam search              & 62.48 & 40.42 & 55.23 \\
                             & Diverse beam search      & 63.83 & 40.62 & 55.76 \\     
\bottomrule
\end{tabular}
}
\caption{\textbf{Oracle {\sc{Rouge-1/2/L}} results} for two decoding methods on all three datasets with a base PEGASUS with 15 generated candidates.}
\label{tab:a_1_1}
\end{table}

As seen in \Cref{tab:a_1_1}, diverse beam search leads to higher oracle values than beam search, motivating our choice to use it.

\begin{table}[h]
\resizebox{\columnwidth}{!}{
\begin{tabular}{lccc}
\toprule
\textbf{Dataset}             
& \textbf{\begin{tabular}[c]{@{}c@{}}R-1 oracle\\ standard dev.\end{tabular}} 
& \textbf{\begin{tabular}[c]{@{}c@{}}R-2 oracle\\ standard dev.\end{tabular}} 
& \textbf{\begin{tabular}[c]{@{}c@{}}R-L oracle\\ standard dev.\end{tabular}} \\
\midrule
\multirow{1}{*}{XSum}        & 14.97 & 18.24 & 16.77 \\
\multirow{1}{*}{Reddit-TIFU} & 14.23 & 12.61 & 13.40 \\
\multirow{1}{*}{SAMSum}      & 15.98 & 21.35 & 18.09 \\
\bottomrule
\end{tabular}
}
\caption{\textbf{Oracle standard deviations} on all three datasets with a base PEGASUS decoded with diverse beam search with 15 generated candidates.}
\label{tab:a_1_2}
\end{table}

\begin{table}[h]
\resizebox{\columnwidth}{!}{
\begin{tabular}{llccc}
\toprule
\textbf{Dataset}             
& \textbf{\begin{tabular}[c]{@{}l@{}}Beam\\ width (m)\end{tabular}} 
& \textbf{\begin{tabular}[c]{@{}c@{}}R-1\\ oracle\end{tabular}} 
& \textbf{\begin{tabular}[c]{@{}c@{}}R-2\\ oracle\end{tabular}} 
& \textbf{\begin{tabular}[c]{@{}c@{}}R-L\\ oracle\end{tabular}} \\
\midrule
\multirow{3}{*}{XSum}        & m=5              & 53.76 & 30.72 & 46.50 \\
                             & m=10             & 56.43 & 33.69 & 49.50 \\
                             & m=15             & 57.82 & 35.28 & 50.75 \\
\midrule
\multirow{3}{*}{Reddit-TIFU} & m=5              & 32.22 & 11.56 & 26.60\\
                             & m=10             & 35.18 & 13.90 & 29.38 \\
                             & m=15             & 36.70 & 15.22 & 30.88 \\
\midrule
\multirow{3}{*}{SAMSum}      & m=5              & 59.03 & 35.21 & 51.25 \\
                             & m=10             & 62.22 & 38.73 & 54.34 \\
                             & m=15             & 63.83 & 40.62 & 55.76 \\
\bottomrule
\end{tabular}
}
\caption{\textbf{Oracle {\sc{Rouge-1/2/L}} results} when varying the beam width on all three datasets with a base PEGASUS with diverse beam search.}
\label{tab:a_1_3}
\end{table}

\newpage
\section{Base Model Hyper-Parameters}
\label{sec:appendix_b}

\begin{table}[h]
\resizebox{\columnwidth}{!}{
\begin{tabular}{lcccccccc}
\toprule
\textbf{Setup}
& \textbf{LR}  
& \textbf{LS}
& \textbf{Optimizer}
& \textbf{BS} 
& \textbf{Epochs} 
& \begin{tabular}[c]{@{}c@{}}\textbf{Max} \\ \textbf{input}\\ \textbf{tokens}\end{tabular} 
& \begin{tabular}[c]{@{}c@{}}\textbf{Max} \\ \textbf{target}\\ \textbf{tokens}\end{tabular} 
& \begin{tabular}[c]{@{}c@{}}\textbf{Eval.}\\ \textbf{frequency}\end{tabular} \\
\midrule
Full-shot   & 1e-4 & 0.1 & Adafactor & 256 & 5 & 512 & 64 & 100 \\
10-shot     & 1e-4 & 0.1 & Adafactor & 5 & 15 & 512 & 64 & 5 \\
100-shot    & 1e-4 & 0.1 & Adafactor & 16 & 15 & 512 & 64 & 10  \\
1000-shot   & 1e-4 & 0.1 & Adafactor & 64 & 15 & 512 & 64 & 10  \\
\bottomrule
\end{tabular}
}
\caption{\textbf{PEGASUS fine-tuning hyper-parameters} in full-shot and few-shot setups used across all three datasets. \textbf{LR} stands for \emph{learning rate}, \textbf{LS} means \emph{label smoothing}, \textbf{BS} is the \emph{effective batch size}, and \textbf{Eval. frequency} represents the number of training batches between two consecutive evaluations of the model during training.}
\label{tab:a_2_1}
\end{table}

\begin{table}[h]
\resizebox{\columnwidth}{!}{
\begin{tabular}{lccccc}
\toprule
\textbf{Dataset} 
& \textbf{\begin{tabular}[c]{@{}c@{}}Max \\ input\\ tokens\end{tabular}} & \textbf{\begin{tabular}[c]{@{}c@{}}Max \\ target\\ tokens\end{tabular}} 
& \textbf{\begin{tabular}[c]{@{}c@{}}Length\\ penalty\end{tabular}} 
& \textbf{\begin{tabular}[c]{@{}c@{}}Repetition\\ penalty\end{tabular}} 
& \textbf{\begin{tabular}[c]{@{}c@{}}Trigram \\ blocking\end{tabular}} \\
\midrule
XSum             & 512 & 64 & 0.8 & 1.0 & Yes \\
Reddit-TIFU      & 512 & 64 & 0.6 & 1.0 & Yes \\
SAMSum           & 512 & 64 & 0.8 & 1.0 & No \\
\bottomrule
\end{tabular}
}
\caption{\textbf{PEGASUS generation hyper-parameters} used to obtain 15 first-stage summary candidates with diverse beam search.}
\label{tab:a_2_2}
\end{table}

\newpage
\section{SummaFusion Hyper-Parameters}
\label{sec:appendix_c}

\begin{table}[h]
\resizebox{\columnwidth}{!}{
\begin{tabular}{lcccccccc}
\toprule
\textbf{Dataset}
& \textbf{LR}  
& \textbf{Optim.}
& \textbf{BS} 
& \textbf{Epochs} 
& \begin{tabular}[c]{@{}c@{}}\textbf{Max} \\ \textbf{input}\\ \textbf{tokens}\end{tabular} 
& \begin{tabular}[c]{@{}c@{}}\textbf{Max} \\ \textbf{tokens}\\ \textbf{per} \\  \textbf{candidate}\end{tabular} 
& \begin{tabular}[c]{@{}c@{}}\textbf{Max} \\ \textbf{target}\\ \textbf{tokens}\end{tabular} 
& \begin{tabular}[c]{@{}c@{}}\textbf{Eval.}\\ \textbf{frequency}\end{tabular} \\
\midrule
XSum         & 2e-5 & Adam & 64 & 5 & 1024 & 34 & 64 & 500 \\
Reddit-TIFU  & 2e-5 & Adam & 64 & 5 & 1024 & 43 & 64 & 100 \\
SAMSum       & 2e-5 & Adam & 64 & 5 & 1024 & 42 & 64 & 150 \\
\bottomrule
\end{tabular}
}
\caption{\textbf{SummaFusion fine-tuning hyper-parameters (full-shot)} for each dataset. \textbf{LR} stands for \emph{learning rate}, \textbf{Optim.} is the \emph{optimizer}, \textbf{BS} is the \emph{effective batch size}, and \textbf{Eval. frequency} represents the number of training batches between two consecutive evaluations of the model during training. We truncate each of the input first-stage candidates to the 95th-percentile value of the summary length distribution on each dataset (represented in the \textbf{Max tokens per candidate} column).}
\label{tab:a_3}
\end{table}

In few-shot SummaFusion fine-tuning, we make the following changes to the values from \Cref{tab:a_2_3}:
\begin{itemize}
    \item 10-shot:
    \begin{itemize}
        \item BS: 4
        \item Epochs: 30
    \end{itemize}
    \item 100-shot:
    \begin{itemize}
        \item BS: 16
        \item Epochs: 30
    \end{itemize}
    \item 10-shot:
    \begin{itemize}
        \item BS: 64
        \item Epochs: 30
    \end{itemize}
\end{itemize}

\section{More Qualitative Examples}
\label{sec:appendix_d}

In the following pages, we show examples of SummaFusion outputs in the same format as \Cref{tab:1} on the other two datasets.

\begin{table*}
\resizebox{0.99\textwidth}{!}{
\begin{tabular}{l}
\toprule
\textbf{Source document}:\\ 
\parbox{1.4\textwidth}{\large 
Hythe Ferry runs between Hythe and Southampton, serviced by a train which runs along a 640m (2,000ft) pier. The presenter lent his support to a community group's aim to take over the management of the pier and train. Its current operator said numbers using the ferry had been falling. Earlier this year, Hythe Ferry Ltd warned staff about possible redundancies, having suffered a "year-on-year decline" in passenger numbers and faced with higher operating costs. More than 9,000 people have signed a petition calling for the service to be saved. Peter King, of the Hythe Hythe Pier Train and Ferry Action Group, said members wanted to create a "viable modern ferry" by a charitable trust taking over the management of the Victorian pier and "relieving" the ferry operators of the costs of maintaining it. He said a redevelopment project could cost £2-3m but a trust would be able to access other sources of finance, including lottery grants. Mr Snow said: "This train is the oldest running pier train anywhere in the world, so I'm campaigning to keep it open. We need to do everything we can to keep this extraordinary piece of our past running."

} \\

\midrule

\textbf{Summary candidates (PEGASUS with diverse beam search)}:\\ 
\parbox{1.4\textwidth}{\large 
1: A campaign to save a ferry service on the Isle of Wight has been backed by BBC Radio Solent presenter Jon Snow. \\
2: BBC Radio 4 presenter Jon Snow has launched a campaign to save a ferry service which is the oldest running in the world. \\
3: TV presenter Jon Snow has launched a campaign to save a ferry service which is the oldest running in the world. \\
4: TV weather presenter Jon Snow has launched a campaign to save a ferry service which is the oldest running pier train in the world. \\
5: The Pier Train is the "oldest running pier train anywhere in the world", according to BBC Radio Solent presenter Jon Snow. \\
6: ITV's This Morning presenter Jeremy Snow has launched a campaign to save a ferry service which is more than 100 years old. \\
7: Snow White has launched a campaign to save a ferry service which is the oldest running in the world. \\
8: BBC Radio 2's Jeremy Snow has launched a campaign to save a historic pier train and ferry service on the Isle of Wight. \\
9: Broadcaster Jon Snow has joined campaigners fighting to save an "extraordinary piece of our past" - a ferry service which is more than 100 years old. \\
10: A campaign to save an "extraordinary piece of our past" has been backed by TV weatherman Jon Snow. \\
11: The X Factor judge Simon Snow has joined campaigners fighting to save a historic pier train and ferry service in Hampshire. \\
12: BBC Radio Solent presenter Jonathan Snow is campaigning to save a "unique" ferry service which is more than 100 years old. \\
13: Comedian Jon Snow is campaigning to save a "unique piece of our past" - a pier train which runs on a ferry. \\
14: Former BBC weatherman Jon Snow is backing an appeal to save the world's oldest pier train service. \\
15: TV presenter Jonathan Snow is backing a campaign to keep the world's oldest pier train running.

}\\

\midrule

\textbf{SummaFusion summary}:\\ 
\parbox{1.4\textwidth}{\large 
\underline{BBC} \underline{Radio} \underline{Solent} presenter \underline{Jonathan} Snow \underline{has} \underline{launched} a \underline{campaign} to \underline{save} a ferry service which is \underline{\emph{believed}} to \emph{be} the oldest running pier train in the world.
} \\

\midrule

\textbf{Ground truth summary}:\\  
\parbox{1.4\textwidth}{\large 
A Hampshire pier and ferry service facing an uncertain future is a "national treasure" which should be saved, television historian Dan Snow has said.
}\\

\bottomrule
\end{tabular}
}
\caption{\textbf{Qualitative sample from the XSum dataset.} Words in the summary from our SummaFusion model which are not in the source document are underlined, and those which are not among any of the first-stage candidates are in italic.}
\label{tab:a_4_1}
\end{table*}


\begin{table*}[h]
\resizebox{0.99\textwidth}{!}{
\begin{tabular}{l}
\toprule
\textbf{Source document}:\\ 
\parbox{1.4\textwidth}{\large this happen yesterday afternoon. i been trying to dual boot mac os and windows on my 
wife's macbook pro. it's a late 2011 model so support from apple is almost nonexistent
which is great when they wanted to charge me chat with them. i convinced them to a 
free chat and learned that apparently my hardware is to out dated to have boot camp 
make a bootable usb. boot camp assistant on this macbook only does cd iso img. \\
...\\ 
i didn't have any blank dvd's laying around because it's 2015! so i made a \\
partition and had my usb in made on my windows desktop. so here is where i fucked up. 1: i never \\
thought to back up any of her files. i knew i had to format the windows partition and then i can \\
move on to installing.  2: trying to do all this pissed and just clicking away. 
3: delete the wrong hard dive part and corrupt the hard drive and have to re format 
the whole computer and lose every file that was saved on her computer since 2011. 
i got yelled at for a good hour. i knew it was my fault but at the same time... 
how in the world have you not backed your things in 4 years!} \\
\midrule
\textbf{Summary candidates (PEGASUS with diverse beam search)}:\\ 
\parbox{1.4\textwidth}{\large 1: got yelled at for 4 years.\\ 
2: i tried to back up my wife's files on her macbook pro and ended up deleting 4 years worth of data.\\ 
3: tried to back up my wife's files on her macbook pro and ended up deleting 4 years worth of data.\\ 
4: got yelled at by a bunch of people because i backed up my wife's files. \\ 
5 i was trying to back up my wife's files on her macbook pro when i accidentally backed it up.\\ 
6: my wife has been using a late 2011 model for 4 years. \\ 
7: 04/21/2019: fired because i backed up my wife's files on her computer since 2011\\ 
8: decided to make a bootable usb on my windows desktop and accidentally backed up my wife's files for 4 years.\\ 
9 got yelled at for 4 hours by a bunch of people because i backed up my wife's files.\\ 
10: got yelled at for 4 years. went to town to get rid of any doubt.\\ 
11: i backed up my wife's files on her macbook pro and now she's going to lose all her files.\\ 
12: i tried to back up my wife's files but ended up deleting 4 years worth of data.\\ 
13: truck driver: i backed up my wife's data for 4 years.\\ 
14: tried to back up my wife's files and ended up deleting her entire computer.\\ 
15: fired because i backed up my wife's data without realizing it. }\\
\midrule
\textbf{SummaFusion summary}:\\ 
\parbox{1.4\textwidth}{\large \underline{tried} to \emph{dual boot} my wife's macbook pro \emph{with boot camp assistant} and \underline{ended} up \underline{deleting} 4 years \underline{worth} of \underline{data}.} \\
\midrule
\textbf{Ground truth summary}:\\ 
\parbox{1.4\textwidth}{\large tried to install windows on macbook and ended up erasing everything without backing up 
and losing 4 years of my wife's work.}\\
\bottomrule
\end{tabular}
}
\caption{\textbf{Qualitative sample from the Reddit TIFU dataset.} The example is the same as in \Cref{tab:1}, but we are including the entire source document. Words in the summary from our SummaFusion model which are not in the source document are underlined, and those which are not among any of the first-stage candidates are in italic.}
\label{tab:a_4_2}
\end{table*}


\begin{table*}
\resizebox{0.99\textwidth}{!}{
\begin{tabular}{l}
\toprule
\textbf{Source document}:\\ 
\parbox{1.4\textwidth}{\large 
Joanne: What are your plans for the holidays? \\
Evelyn: Nothing. I’ll stay at home and rest. \\
Joanne: You must be exhausted after the past few weeks \\
Evelyn: It’s been hectic \\
Joanne: I’m going back home. \\
Evelyn: To France? \\
Joanne: Yes. Not that I want to go… \\
Evelyn: Why? You always liked spending Christmas with your family. \\
Joanne: I did. But my parents separated a few months ago \\
Joanne: It is still pretty tense… \\
Evelyn: I’m sorry to hear that \\
Joanne: My dad left my mum for his secretary \\
Joanne: Such a cliché  \\
Joanne: My mum is devastated \\
Joanne: So I’m basically going to cheer her up \\
Joanne: It’s really hard for her now \\
Joanne: For me it’s also not easy \\
Evelyn: I can imagine! \\
Evelyn: If you want to bring your mum over here we could spend Christmas together. \\
Joanne: Thanks, that’s really sweet. But I don’t think she’s in a condition for that. She’s been very depressive lately.
} \\

\midrule

\textbf{Summary candidates (PEGASUS with diverse beam search)}:\\ 
\parbox{1.4\textwidth}{\large 
1: Evelyn will stay at home and rest for the holidays. Joanne is going back home. Her parents separated a few months ago. Joanne's mum is depressed. Evelyn suggests that they \\
2: Joanne is going back home for Christmas. Her parents separated a few months ago. Joanne's mother is depressed. Evelyn suggests that they could spend Christmas together. \\
3: Evelyn will not go to France for Christmas. Joanne's parents separated a few months ago. Joanne's mum is devastated. Evelyn suggests that Joanne bring her mum over to France \\
4: Joanne wants to spend Christmas with her family, but her parents separated a few months ago. Evelyn wants to spend Christmas with Joanne's mother. \\
5: Joanne is going back home for Christmas. Her parents separated a few months ago. Joanne's mother is depressed. Evelyn suggests that they could spend Christmas together. Joanne's mother is \\
6: Joanne wants to spend Christmas with her family, but her parents separated a few months ago. Evelyn wants to spend Christmas with Joanne's mother. She's depressed lately. \\
7: Joanne's parents separated a few months ago. She's going home for Christmas. Evelyn suggests that she could bring her mother over to France. \\
8: Joanne's parents separated a few months ago. She's going home for Christmas. Evelyn suggests that she could bring her mother over to France. Joanne's mum is depressed lately. \\
9: Evelyn won't go to France for Christmas. Joanne's parents separated a few months ago. Joanne's mum is devastated. Evelyn suggests that they could spend Christmas together. \\
10: Evelyn has been busy for the holidays. Joanne is going back home. Her parents separated a few months ago. Joanne's mum is depressed. Evelyn suggests that they could spend Christmas \\
11: Joanne's parents separated a few months ago and she doesn't want to go to France for Christmas. Evelyn suggests that she could bring her mother over to France for Christmas. \\
12: Joanne and Evelyn are going back home to spend Christmas with their families. Joanne's parents separated a few months ago. Joanne's mum is depressed. Evelyn suggests that they could \\
13: Joanne is going home for Christmas. Her parents separated a few months ago. Joanne's mother is depressed and she doesn't want to spend Christmas with her family. \\
14: Evelyn doesn't have any plans for the holidays. Joanne's parents separated a few months ago. Joanne's mum is devastated. Evelyn suggests that they could spend Christmas together. \\
15: It has been hectic for Evelyn and she will not go to France for Christmas. Joanne's parents separated a few months ago and she will cheer her mother up.
}\\

\midrule

\textbf{SummaFusion summary}:\\ 
\parbox{1.4\textwidth}{\large 
Evelyn \underline{will} stay at home for the holidays. Joanne is going back home to France. She \emph{always liked spending} Christmas with her family, but her parents separated a few months ago and her \underline{mother} is \underline{depressed}. Evelyn \underline{\emph{offers}} to bring her \underline{mother} over for Christmas.
} \\

\midrule

\textbf{Ground truth summary}:\\ 
\parbox{1.4\textwidth}{\large 
Joanne is going to go back home to France for the holidays. She's going to cheer her mum up because her parents separated a few months ago. Evelyn offers Joanne to spend Christmas together if she brings her mum over here.
}\\

\bottomrule
\end{tabular}
}
\caption{\textbf{Qualitative sample from the SAMSum dataset.} Words in the summary from our SummaFusion model which are not in the source document are underlined, and those which are not among any of the first-stage candidates are in italic.}
\label{tab:a_4_3}
\end{table*}

\end{document}